\documentclass[sigconf]{acmart}
\AtBeginDocument{%
  }

\copyrightyear{2022} 
\acmYear{2022} 
\setcopyright{rightsretained} 
\acmConference[ICPP '22]{51st International Conference on Parallel Processing}{August 29-September 1, 2022}{Bordeaux, France}
\acmBooktitle{51st International Conference on Parallel Processing (ICPP '22), August 29-September 1, 2022, Bordeaux, France}\acmDOI{10.1145/3545008.3545020}
\acmISBN{978-1-4503-9733-9/22/08}

\usepackage{bbold}
\usepackage{physics}
\usepackage[ruled, vlined, linesnumbered]{algorithm2e}

\begin{document}

\title{HARL: Hierarchical Adaptive Reinforcement Learning Based Auto Scheduler for Neural Networks}

\author{Zining Zhang$^{1, 2}$}
\email{zzn@nus.edu.sg}
\affiliation{%
  \institution{$^1$School of Computing, National University of Singapore}
  \institution{$^2$NUS Centre for Trusted Internet and Community}
  \country{Singapore}}
\author{Bingsheng He}
\email{hebs@comp.nus.edu.sg}
\affiliation{%
  \institution{School of Computing, National University of Singapore}
  \country{Singapore}}
\author{Zhenjie Zhang}
\email{zhenjie.zhang@neuron.sg}
\affiliation{%
  \institution{Neuron Mobility Pte. Ltd.}
  \country{Singapore}}
  
\begin{abstract}
    To efficiently perform inference with neural networks, the underlying tensor programs require sufficient tuning efforts before being deployed into production environments. Usually, enormous tensor program candidates need to be sufficiently explored to find the one with the best performance. This is necessary to make the neural network products meet the high demand of real-world applications such as natural language processing, auto-driving, etc. Auto-schedulers are being developed to avoid the need for human intervention. However, due to the gigantic search space and lack of intelligent search guidance, current auto-schedulers require hours to days of tuning time to find the best-performing tensor program for the entire neural network.
    
    In this paper, we propose HARL, a reinforcement learning (RL) based auto-scheduler specifically designed for efficient tensor program exploration. HARL uses a hierarchical RL architecture in which learning-based decisions are made at all different levels of search granularity. It also automatically adjusts exploration configurations in real-time for faster performance convergence.
    As a result, HARL improves the tensor operator performance by 22\% and the search speed by 4.3x compared to the state-of-the-art auto-scheduler.
    Inference performance and search speed are also significantly improved on end-to-end neural networks.
    
\end{abstract}



\maketitle

\section{Introduction}
Deep neural networks (DNNs) with high performance requirements have become ubiquitous in AI applications such as auto-driving~\cite{cordts2016cityscapes}, real-time speech recognition~\cite{hannun2014deep}, language translation~\cite{devlin2018bert}, etc.
To achieve the performance requirements under the circumstances of rapidly evolving neural networks and hardware platforms, considerable engineering efforts are frequently required to tune the underlying vendor-provided libraries like oneDNN~\cite{oneapi} and cuDNN~\cite{chetlur2014cudnn} for neural models.
For example, even a single $1024 \times 1024 \times 1024$ matrix multiplication operator contains $\sim$180 millions of possible implementations for exploration.
To liberate engineers and researchers from these labor-intensive tuning tasks, auto-schedulers~\cite{chen2018learning, adams2019learning, zheng2020ansor} are developed for automatic tuning of tensor programs.
They efficiently search the large search space with the help of statistical optimization methods such as simulated annealing~\cite{van1987simulated} and evolutionary search~\cite{mitchell1998introduction}.

However, statistical approaches do not consider the varying performance distributions under different tensor programs, hardware platforms, and parameter sets, leading to suboptimal search efficiency and effectiveness.
For example, AutoTVM~\cite{chen2018learning} applies a simulated annealing approach to select the parameter tuning directions throughout the search.
Although this method is able to approximate the best-performed tensor program theoretically,
 simulated annealing assumes heuristic probabilities of the acceptable sets of parameters, which makes the search inefficient.

There are learning-based auto-schedulers proposed to replace the statistical search process.
For example, Flextensor~\cite{zheng2020flextensor} and Chameleon~\cite{ahn2020chameleon} utilize the decision power of reinforcement learning (RL).
They greatly outperform AutoTVM in both inference performance and search speed.
However, a lately developed statistical auto-scheduler Ansor~\cite{zheng2020ansor} utilizes the more advanced evolutionary search which keep evolving the tensor program candidates in a large search space with a wide coverage on tuning options.
Both Flextensor and Chameleon are beaten by Ansor on the search speed and the performance of a wide range of tensor operators, even though Ansor assumes fixed and heuristic probabilities while tuning the parameters. Additionally, since Ansor is capable of optimizing combined tensor operators, it is easy to be applied in end-to-end neural network optimization.
We observe that there are following challenges to align RL-based auto-schedulers with Ansor:
1) \emph{Lack of suppport on end-to-end neural networks optimization}:
reinforcement learning requires grid-like search space which suits the operator-level parameter tuning well, but is unable to solve the complicated combinations of multiple operators.
2) \emph{Wasteful search iterations on poor search paths}:
reinforcement learning normally requires a termination condition to indicate that the endpoint has been reached. However, in tensor program parameter search, there is no endpoint, i.e. every set of parameters is available for performing the next action. Thus, RL methods tune a whole batch of parameters for a predefined uniform number of steps, without considering the performance potentials case-by-case.

In this paper, we propose HARL, a hierarchical and adaptive RL-based auto-scheduler that eliminates the heuristic and greedy processes in tensor program search.
HARL is capable of tuning the whole neural networks by applying a hierarchical RL decision-making process using separate RL models for different levels of search hierarchy: subgraphs, subgraph structures (sketches), and low-level operator parameters.
HARL also manages the exploration lengths adaptively by allocating more tuning time slots to potentially well-performing programs in a finer granularity than a whole batch.
With the help of both the hierarchical and adaptive learning-based searching schemes, HARL is able to exploit the power of RL algorithms and out-performs Ansor in both the tuning speed and the performance of the final tensor program.

In summary, the contribution of this paper are as follows:
\begin{itemize}
    \item A \emph{learning-based} search method for exploring high performance tensor programs
    \item A \emph{hierarchical} search scheme to improve efficiency at all search granularities using RL algorithms
    \item An \emph{adaptive-stopping} search process for wisely pruning the less efficient search paths
    \item A thorough evaluation of the above system, which outperforms prevailing auto-schedulers~\cite{chen2018learning,adams2019learning,zheng2020ansor} in both tuning speed and program performance on CPU/GPU platforms.
\end{itemize}

Experiments are performed on different tensor computations and on different underlying hardware platforms.
The results show that
for the tensor operators, HARL is able to achieve the same performance up to 4.3x faster than Ansor. 
Furthermore, the performance of the final tensor program outperforms that of Ansor by 22\%.
For end-to-end neural networks, the search speed and the inference performance improvement are up to 2.2x and 8\%, respectively.

The paper is structured as follows:
Section~\ref{sec:bg} explains the background of current auto-schedulers and our motivations.
Section~\ref{sec:overview} gives an overview of the HARL system.
Section~\ref{sec:hier_search} and Section~\ref{sec:adaptive_search} elaborate on the details of the main contributions.
Section~\ref{sec:eval} shows the evaluation setup and analysis on different tensor computations and different hardware platforms.
Section~\ref{sec:conclusion} summarizes this work.

\section{Background \& Related Work}
\label{sec:bg}

\subsection{Problems of Current Auto-Schedulers}

\textbf{Statistical Auto-Schedulers}:
Statistical auto-schedulers utilize the statistical optimization approaches to explore the tenosr programs.
Although the search process is heuristic and greedy, they are the prevailing techniques used in research and industry.
AutoTVM~\cite{chen2018learning}, for example, limits the search space by requiring user-provided templates. It uses simulated annealing~\cite{van1987simulated} to search the optimal parameters under the specified template.
However, the predefined templates limit the possibilities of different combinations of multiple operators and thus lead to suboptimal programs.
On the other hand, Halide auto-scheduler~\cite{adams2019learning} solves the problem by defining the search space as a sequence of choices of the program construction.
To explore this search space, Halide auto-scheduler builds a decision tree that helps sequentially fill the parameters of computational DAG (Directed Acyclic Graph) nodes of the tensor program.
The branches of the decision trees are pruned using a cost model that predicts the final performance on the partially constructed programs.
This process is applied with general rules and therefore does not need any template.
However, the performance prediction on incomplete programs leads to inaccurate predictions.
Ansor~\cite{zheng2020ansor} eliminates the drawbacks of the previous two methods: it is template-free and predicts the schedule performance based on complete programs.
It generates high-level program sketches based on the abstract definition of the tensor computation, and then explores and fills the low-level parameters using evolutionary search~\cite{mitchell1998introduction}.
The sketches generated by Ansor eliminate the need for templates. 
The low-level parameters are fully filled during evolutionary search, thus the performance estimations are based on complete programs. With these fine properties and the effectiveness brought by evolutionary search, Ansor is the state-of-the-art auto-scheduler.

\textbf{RL-based Auto-Schedulers}: 
Flextensor~\cite{zheng2020flextensor} and Chameleon~\cite{ahn2020chameleon} improve the search efficiency using RL algorithms.
Flextensor is able to cover the cases of different tensor operators with the help of a general template, but fails to handle the optimization of end-to-end neural networks.
Chameleon is another auto-tuner which improves the performance of AutoTVM by replacing the simulated annealing with a reinforcement learning model.
It also adaptively manages the initial schedule sampling, which is originally uniform sampling in AutoTVM.
However, Chaemeleon follows the template-based optimization process of AutoTVM, thus the performance of the outcome depends on the quality of the user-provided templates.
Note that both Flextensor and Chaemeleon ignore the importance of the time-allocation for different tuning paths, and thus are out-performed by Ansor. Bansor~\cite{gao2021bansor} is an incremental improvement over Ansor which focuses only on modifying the subgraph and sketch selections as bandit-based.

\begin{table*}
  \caption{System Comparison: Ansor vs. Flextensor vs. HARL. HARL uses either learning-based or adaptive methods on all the 4 search processes. MAB stands for multi-armed bandit~\cite{vermorel2005multi}.}
  \label{tab:assum_cmp}
  \begin{tabular}{cp{3cm}p{3cm}p{3cm}p{5cm}}
    \toprule
    & Subgraph Selection \newline $\pi_t(n)$ & Sketch Selection \newline $\pi^n_t(u)$ & Schedule Selection \newline $\pi^{n, u}_t(s_t|s_{t-1})$ & Schedule Tracks Time-Allocation \newline $\{T^n_i\}_{1\le i \le I}$ \\
    \midrule
    Ansor~\cite{zheng2020ansor} & {Greedy Selection}$\ $ & {Uniform Distribution}$\ $ & {Uniform Distribution}$\ $ & {Greedy Allocation}$\ $ \\
    Flextensor~\cite{zheng2020flextensor} & {Not Supported}$\ $ & {Fixed Sketch}$\ $ & RL Agent & {Uniform Allocation}$\ $ \\
    HARL & MAB RL & MAB RL & RL Actor Network & Estimation on Future Performance \\
  \bottomrule
\end{tabular}
\end{table*}

\subsection{Problem Formulation \& Motivation}
\label{ssec:problem}
For a search-based tensor program scheduler, the search problem can be treated as finding the scheule $S$ with the best inference performance which is measured by the execution time $f(S)$.
Since neural networks are mostly modularized, the computational graph of the network is first splitted into $N$ subgraphs that are sequentially executed.
The optimization target is thus approximated by the estimation $f(S) \approx \sum_{n=1}^{N} w_n \times g_n$, where $w_n$ and $g_n$ are the number of appearances and the execution time of the subgraph $n$, respectively.
At each tuning step $t$, the auto-scheduler chooses a subgraph $n$ with probability $\pi_t(n)$ for optimization.
Under this subgraph selection distribution, the total tuning time-allocation for each subgraph $n$ is denoted as $T^n$, and $\sum_{n=1}^{N} T^n = T$, where $T$ is the total time-slots for the neural network optimization.

Subgraph $n$ is assumed to contain one or more tensor operators, and thus has different structures (or \emph{sketches}) on operator combinations.
Suppose that there are a total of $U_n$ sketches for the subgraph $n$.
Sketch $u$ is selected as the skeleton of the subsequent parameter search.
The distribution of sketch selection at step $t$ on subgraph $n$ is represented as $\pi_t^n(u)$.

For the selected sketch, the low-level parameters are explored by first selecting an initial schedule of the parameters set, and then modifying and measuring the schedule step-by-step.
Suppose that there are a total of $I$ initial schedules sampled, since each of them is modified and measured independently, we call each search path starting from an initial schedule as a \emph{schedule track} $i$.
The schedule modification process follows the Markov Decision Process (MDP)~\cite{otterlo2012reinforcement}.
$s^n_{i, t}$ is defined as the schedule when optimizing subgraph $n$ at step $t$ of schedule track $i$.
We define $\pi_t^{n, u}(s^n_{i, t}|s^n_{i, t-1})$ as the probability of transition from schedule $s^n_{i, t-1}$ to $s^n_{i, t}$. 
It is also written as $\pi_t^{n, u}(s_{t}|s_{t-1})$ for brevity.
$T^n_i$ is defined as the allocated slots for schedule track $i$ under the optimization of subgraph $n$, such that $\sum_{i=1}^{I} T^n_i = T^n$.
The target of the subgraph parameter tuning for each schedule track $i$ is to find a series of schedules $s^n_{i, 1}, s^n_{i, 2}, ... s^n_{i, T^n}$ with maximum final performance on $s^n_{i, T^n}$.

\begin{figure}[htb]

\begin{minipage}[t]{1\linewidth}
  \vspace{0pt}
  \centering
  \centerline{\includegraphics[width=8cm]{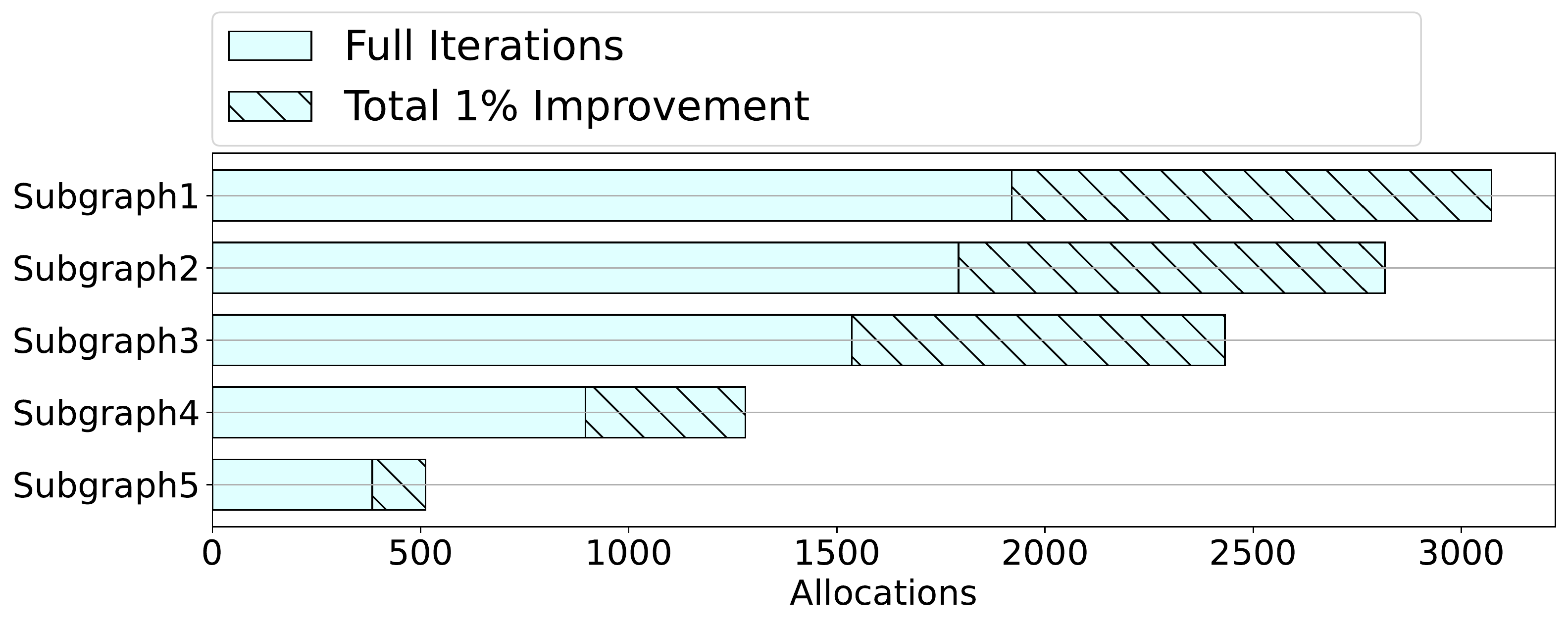}}
  \begin{center}{(a) Time-allocations on top-5 most time-consuming subgraphs of BERT. The sketched bars are the number of trials allocated for last 1\% improvement.}\end{center}\medskip
\end{minipage}
\begin{minipage}[t]{0.48\linewidth}
  \vspace{0pt}
  \centering
  \centerline{\includegraphics[width=4.1cm]{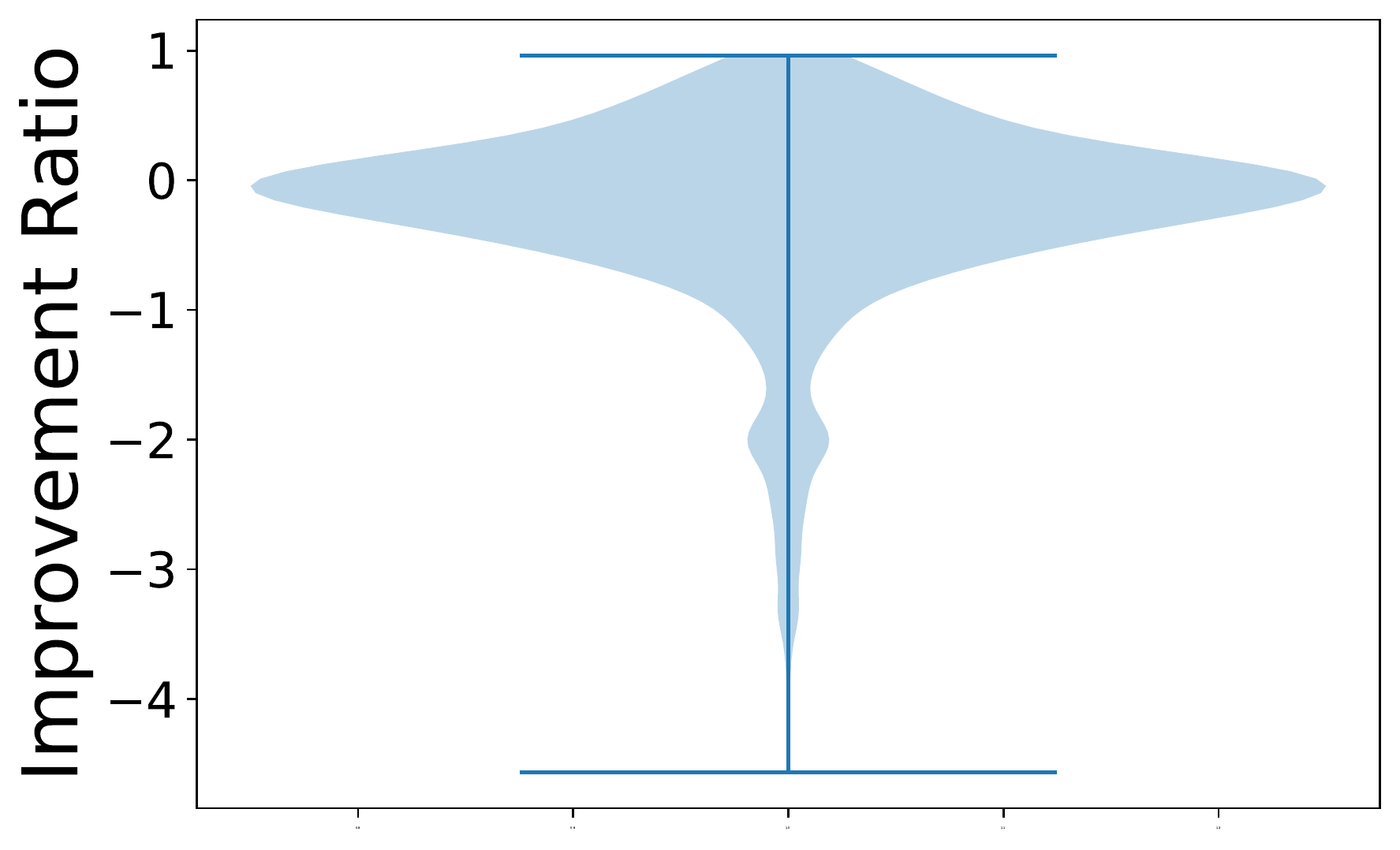}}
  \begin{center}{(b) Performance improvement distribution. \emph{Improvement Ratio} is the performance improvement from original schedule to a next schedule relative to the performance of the original schedule.}\end{center}\medskip
\end{minipage}
\begin{minipage}[t]{0.48\linewidth}
  \vspace{0pt}
  \centering
  \centerline{\includegraphics[width=4.1cm]{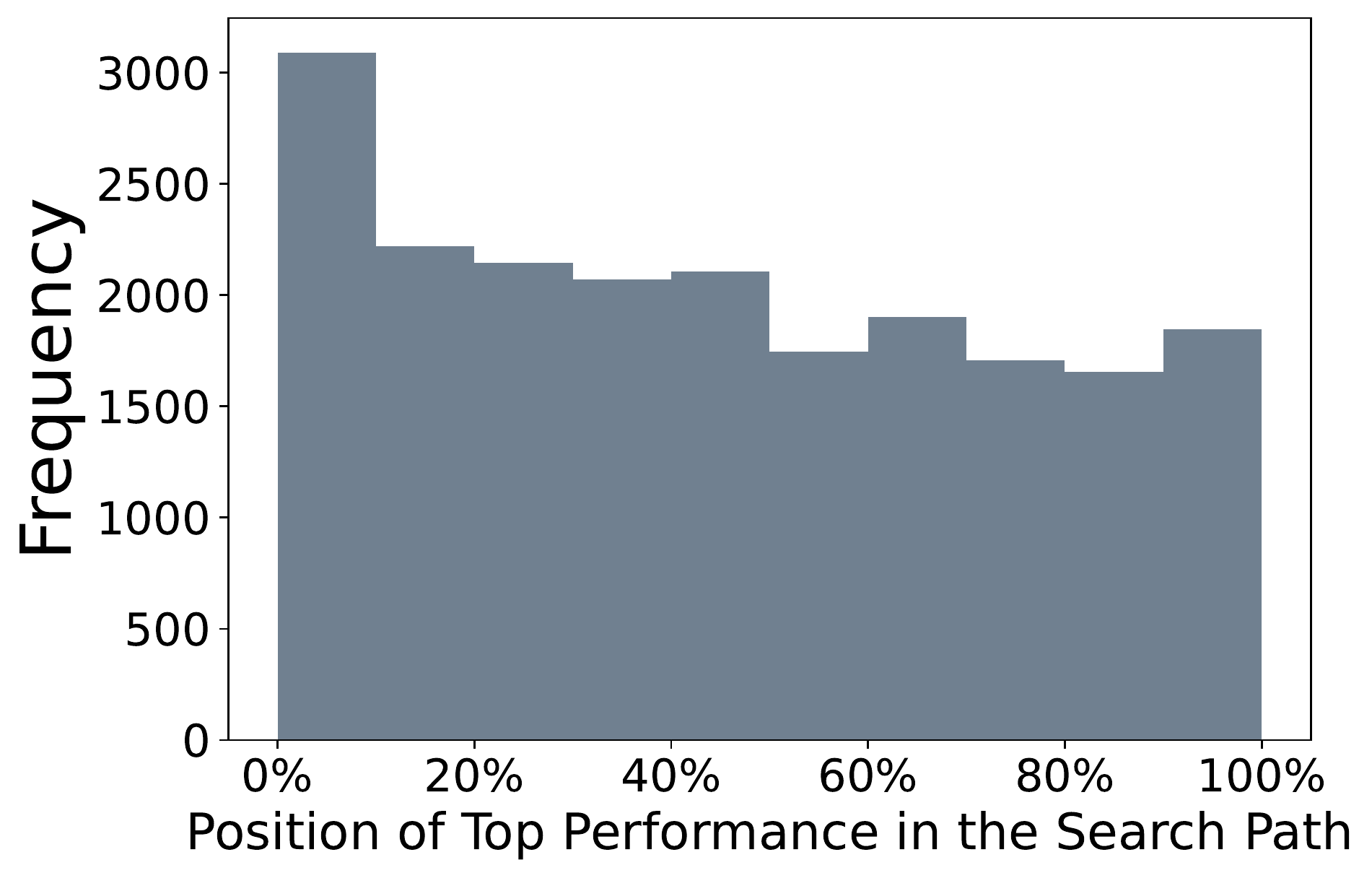}}
  \begin{center}{(c) Histogram of search path efficiency on Flextensor. X-axis represents the position of the best-performed schedule on the search path divided by the path length. Y-axis is the frequency of these relative positions collected during auto-tuning.}\end{center}\medskip
\end{minipage}
\caption{Observations made on current auto-schedulers.}

\label{fig:observations}
\end{figure}

Normally, different auto-schedulers have different assumptions on the graph selection distribution $\pi_t(n)$, sketch selection distribution $\pi_t^n(u)$, schedule selection distribution $\pi_t^{n,u}(s_t|s_{t-1})$, and schedule track slots $T^n_i$ allocations.
We have performed experiments on the current statistical and RL based search methods, and there are following observations that indicate the short-comings of the above distributions assumed in the current auto-schedulers:

\emph{Observation 1: Greedy and heuristic subgraph/sketch/schedule selection distributions $\pi(\cdot)$ waste the optimization iterations significantly.}
When applying the greedy selection process of subgraphs $\pi_t(n)$ used in Ansor for neural network optimizations, an excessive number of trials are allocated for the most time-consuming subgraphs. As shown in Figure~\ref{fig:observations}(a), the sketched bars are the part of allocated trials that only improve overall performance by 1\%.
Over 35\% of the total trials are spent on this tiny improvement.
We observe that, the top 3 subgraphs have no improvement. Only \emph{Subgraph5} contributes to 50\% of the 1\% improvement.
The greedy allocation process only considers the previous improvements of each subgraph and its contribution to the total execution time, thus it ignores the importance of \emph{exploration} as RL does.
Experiments in Section~\ref{ssec:exp_nn} shows that by allocating more time-slots to \emph{Subgraph5} the total performance is able to improve more.
On the other hand, we also performs experiments using the uniform schedule selection $\pi_t^{n,u}(s_t|s_{t-1})$ of Ansor.
In Figure~\ref{fig:observations}(b), 200 programs are tested with uniformly selecting the next schedules for 20 trials. The improvements from these trials are shown in a violin plot.
Most of the improvements are around 0. Although the schedule selection process of Ansor is powered by the evolutionary search, this uniform selecting still harms the search efficiency. Thus, a learning-based $\pi^{n,u}_t(s_t|s_{t-1})$ is desired.

\emph{Observation 2: Uniform schedule track allocations $\{T^n_i\}_{1\le i \le I}$ result in high ratio of useless search steps.}
When using fixed-length schedule search on a batch of parameters, some search paths may already reach the optimal points at the early steps, thus the following steps do not contribute.
In Figure~\ref{fig:observations}(c), various GEMM operations are performed on Flextensor which uses fixed-length search. The histogram shows that most of the search paths find the best parameters within the first 40\% of the total search steps. This verifies the search inefficiency mentioned above.

With the above two observations, a search method with adaptive $\pi_t(n), \pi_t^n(u), \pi_t^{n, u}(s_t|s_{t-1}), \{T_i^n\}_{1 \le i \le I}$ distributions is required for a more efficient neural network optimization process.
With this motivation, HARL is developed. As shown in Table~\ref{tab:assum_cmp}, HARL applies learning-based search algorithms that adaptively learns the time-varying sampling distribution of subgraphs and their sketches. 
It explores the parameter search space with an RL agent.
The optimization time slots for subgraphs and schedule tracks are adaptively controlled for better search efficiency.
Compared with HARL, Ansor uses greedy or heuristic distributions in these optimization processes.
While Flextensor applies RL-based schedule selection process, it does not support optimizing combinations of tensor operators with the help of subgraph and sketch selections, and it simply uses fixed-length uniform allocations on different schedule tracks.

\begin{figure*}[h]
  \centering
  \includegraphics[width=0.7\linewidth]{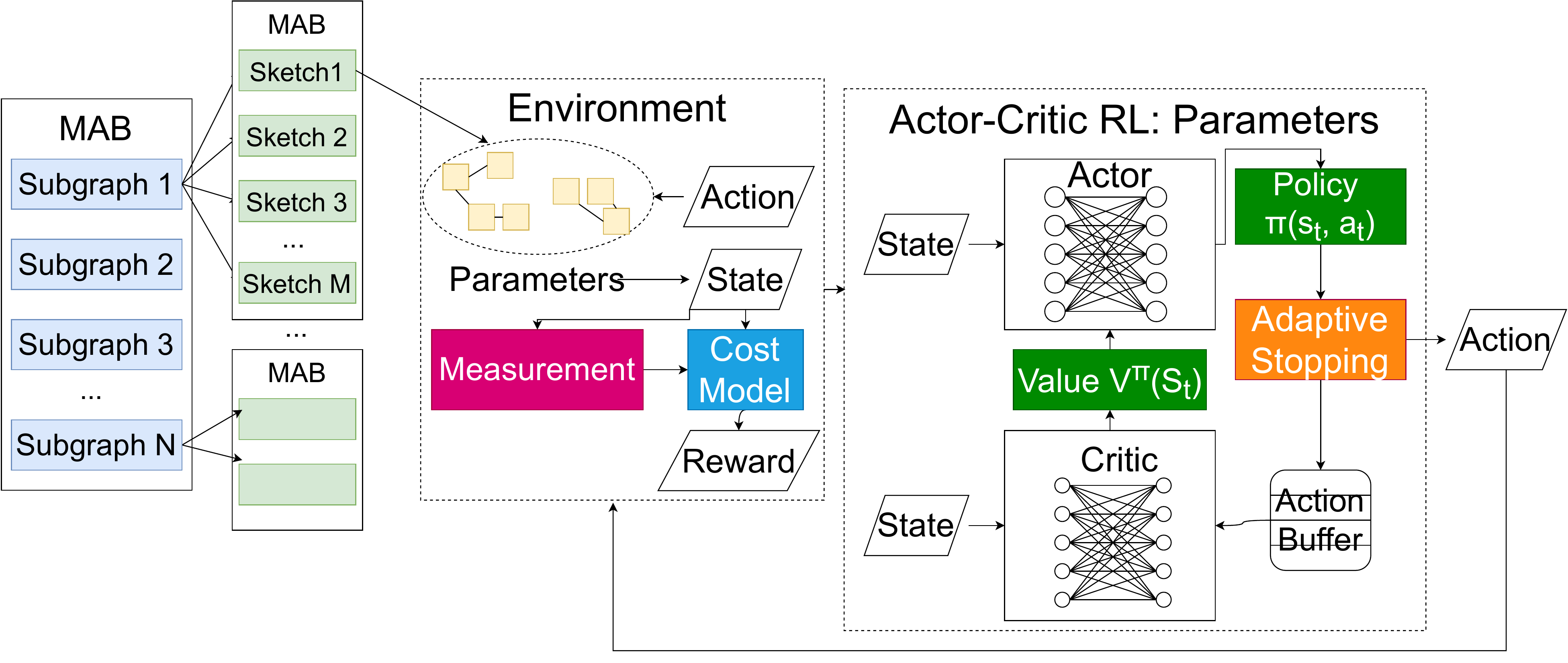}
  \caption{HARL overview}
  \label{fig:overview}
  \Description{overview of HARL}
\end{figure*}

\section{System Overview}
\label{sec:overview}

\subsection{Search Hierarchy}
HARL adopts a hierarchical search scheme as described in Section~\ref{ssec:problem}.
First, the end-to-end neural network optimization problem is divided into \textit{subgraph} optimization problems.
Each subgraph $n$ may have multiple computational structures or \textit{sketches}.
The sketches of the subgraph are automatically generated based on the computational definition of the subgraph. The generation rules are the same as those of Ansor~\cite{zheng2020ansor}, and are shown in Table~\ref{tab:sketch_rules}.
\emph{Tiling} determines the loop structures (tiling sizes and loop orders) of the tensor programs.
\emph{Inline} configures the locations of the computations.
\emph{rfactor}~\cite{suriana2017parallel} is to apply parallelism on the reduction dimensions.
\emph{Cache Write} adds an additional operation that caches the writes to the output buffer.
Applying these rules differently results in different sketches of the optimizing tensor program.

Next, HARL tunes the parameters on a specific sketch $u$ of the target subgraph $n$. There are a couple of tuning knobs for the auto-scheduler to modify to transit to the next state. 
The tuning knobs, which are also known as \textit{modification types} contain computation-location modification, tile-size modification, parallelism modification, and annotation modification.
Section~\ref{ssec:search_space} describes all the \textit{modification types} used in HARL in details.
Then, the auto-scheduler chooses a specific set of new paraemters of the selected modification type as instructed by the search algorithm.
These new parameters modified from schedule $s_{t-1}$ are applied to generate the new schedule $s_t$ for the next exploration step.

In summary, combined with the problem definition specified in Section~\ref{ssec:problem}, there are four different levels of search hierarchy that are controlled by the search algorithm:
\textit{subgraph} (a.k.a. task) $n$ with sampling distribution $\pi_t(n)$, \textit{sketch} $u$ with sampling distribution $\pi^n_t(u)$, \textit{modification type}, and \textit{parameter tuning} transition distribution $\pi_t^{n, u}(s^n_t|s^n_{t-1})$.

\begin{table}
  \caption{Sketch Generation Rules Adopted from Ansor~\cite{zheng2020ansor}}
  \label{tab:sketch_rules}
  \begin{tabular}{cp{5.5cm}}
    \toprule
    Rule Name&Description\\
    \midrule
    Skip & Skip any modification if not able to inline \\
    Inline & Inline the function if it's possible \\
    Tiling& Tile the loops if the function has data reuse\\
    Tiling with Fusion& Tile the loops and fuse with the consumer if has data reuse\\
    Cache Write& Cache the output if has data reuse but without any consumers\\
    rfactor& Perform reduction factorization if has reduction parallelism\\
  \bottomrule
\end{tabular}
\end{table}

\subsection{HARL Structure}
HARL applies RL modeling to all levels of the search hierarchy.
Figure~\ref{fig:overview} shows the overview of the HARL system.
It applies separate RL models for different search granularities.

For \textit{high-level} subgraph and sketch selections, HARL assumes non-stationary distribution of the rewards.
As the search goes on, if a subgraph is optimized for most of the time, it is unlikely to keep the same distribution of improvements.
This time-varying reward distribution indicates a time-varying subgraph sampling distribution $\pi_t(n)$.
The sampling distribution affects the subgraph time-allocations $\{T^1, T^2, ..., T^N\}$.
For the sketch selection distribution, it also changes over time.
In this case, both the subgraph selection and sketch selection phases are modeled as a non-stationary multi-armed bandit problem (MAB)~\cite{vermorel2005multi}. 
This is different from Ansor's approach, where the subgraph sampling distributions $\pi_t(n)$ is modeled as a greedy process with deterministic selections, and $\pi_t^n(u)$ is treated as time-independent uniform distribution. More details are explained in Section~\ref{ssec:bandit}.

For the \textit{low-level} modification types and parameter tuning, the MAB model is no longer applicable, since there are an extraodinary number of parameters for each sketch.
In this case, the search space of parameters can be constructed as connected graphs as shown in the RL environment in Figure~\ref{fig:overview}.
Each step of parameter tuning is thus modeled as an edge connecting one schedule setting to another.
This edge is treated as an action generated by a Markov decision process (MDP) $\pi^{n, u}_t(s_t|s_{t-1})$, thus the parameter search problem is solvable by RL algorithms like DQN~\cite{mnih2013playing}, actor-critic method~\cite{mnih2016asynchronous}, etc.
The construction of this search space is explained in more details in Section~\ref{ssec:search_space}.

Compared with the ordinary actor-critic RL algorithm, there are two additional modules in HARL: a \textbf{cost model} and an \textbf{adaptive stopping} module.
To avoid a large number of time-consuming measurements, we use a light-weight cost model to predict the actual performance of each schedule.
This cost model prunes the schedules with low prediction scores and learns on the fly from the actural measurements.
It is also used as the reward function in the RL model.
The adaptive stopping module is used to adaptively allocates the times slots $\{T^n_1, T^n_2, ..., T^n_I\}$ for schedule tracks $i \in [1, I]$.
This track-wise fine-grained length-control module helps to explore more on schedules with better potentials.
The details are presented in Section~\ref{sec:adaptive_search}.

\section{Hierarchical Searching}
\label{sec:hier_search}
\subsection{Bandit-based RL for High Level Decisions}
\label{ssec:bandit}
As described in Section~\ref{sec:overview}, the decision-making process of schedule search goes through four levels of hierarchy: \textit{subgraph} selection, \textit{sketch} selection, \textit{modification type} selection and the \textit{modification parameter} selection.
We observe that high-level decisions usually have a fixed and limited number of choices.
For example, in a BERT~\cite{devlin2018bert} model, the number of distinct subgraphs is 10, and that of ResNet-50~\cite{he2016deep} is 24.
For a \textit{matrix multiplication} subgraph, the number of sketches is 3.
On the other hand, the optimization outcome for each step of these subgraph/sketch selections is with a non-deterministic performance improvement.

Based on these observations, the subgraph selection and the sketch selection process match the MAB problem~\cite{vermorel2005multi}.
Additionally, the two MAB problems are non-stationary as the performance of different subgraphs and sketches continues to improve over the course of optimization.
To describe our solution to the MAB problems, the following terms are introduced:
In order to trace the history of the performance of the visited schedules, the normalized maximal performance until each step is recorded in a list $\{X_t\}_{1 \le t < T}$.
The list of selected subgraphs and the list of selected sketches are recorded in $\{O_t^N\}_{1\le t < T}$ and $\{O_t^U\}_{1 \le t < T}$, respectively.
To optimize the two non-stationary MAB problems, we use Sliding-Window Upper-Confidence Bound (SW-UCB)~\cite{garivier2008upper}:
\begin{equation}
\label{eq:sw-ucb}
O_t = \operatorname*{argmax}_a \Bigg(
    Q_t(\tau, a) + 
    c\sqrt{\frac{\ln(\min(t, \tau))}{\mathcal{N}_t(\tau, a)}} 
\Bigg)
\end{equation}

In this definition, $a$ represents the action, and in the case of HARL, it is the subgraph/sketch that is considered.
$c$ is a hyperparameter that determines the importance of exploration.
$\tau$ is the window size.
$t$ is the current step, and $\mathcal{N}_t(\tau, a)$ is the number of allocated trials/steps on the subgraph/sketch $a$, from step $t-\tau+1$ to step $t$.

$Q_t(\tau, a)$ is the average reward evaluated for action $a$ at the step $t$.
For sketch selection, it is simply defined as the average performance in the window with size $\tau$:
\begin{equation}
\label{eq:qts}
Q_t^U(\tau, a) = \frac{1}{\mathcal{N}_t(\tau, a)} \sum_{t' = t-\tau+1}^{t} X_{t'} \mathbb{1}_{\{O_{t'}^U = a\}}
\end{equation}
where $a$ specifically represents the action of \emph{sketch} selection.

For \emph{subgraph} selection, the performance is different for each subgraph, thus instead of using the absolute performance, 
HARL directly uses the gradient estimation formula in~\cite{zheng2020ansor} as the reward function for prediciting the performance \emph{improvement} when selecting a subgraph:
\begin{equation}
\label{eq:ansor_task}
\begin{split}
R_t(a) = &\Bigg| \pdv{f}{g_a} \Big( \alpha \frac{g_a(t_a) - g_a(t_a-\Delta t)}{\Delta t} + \\
&(1-\alpha) \min(-\frac{g_a(t_a)}{t_a}, \beta \frac{B_a}{\max_{a' \in M(a)}P_{a'}} - g_a(t_a)) \Big) \Bigg| \\
\end{split}    
\end{equation}
This gradient estimation formula predicts the benefit from optimizing the subgraph based on both the performance improvement from the previous steps (the first term), as well as the execution-time contribution and the similar subgraphs (the second term). Please refer to the original paper~\cite{zheng2020ansor} for details.
Finally, the $Q_t(\tau, a)$ for \emph{subgraph} selection is defined as:
\begin{equation}
\label{eq:qtg}
Q_t^N(\tau, a) = \frac{1}{\mathcal{N}_t(\tau, a)} \sum_{t' = t-\tau+1}^{t} R_{t'}(a) \mathbb{1}_{\{O_{t'}^N = a\}}
\end{equation}

By utilizing the SW-UCB for solving the non-stationary MAB subgraph/sketch selection problem, the search process starts to \emph{explore} the search space instead of greedily selecting the action solely based on the previous data or estimations.
The exploration importance is controled by the constant $c$ in Equation~\ref{eq:sw-ucb}. Its following term gives more oppurtunities to the actions that are seldomly explored in the previous rounds.

\subsection{Parameter Search Space}
\label{ssec:search_space}

\begin{table*}
  \caption{Types of Parameter Modifications}
  \label{tab:mod_kind}
  \begin{tabular}{ccp{7cm}}
    \toprule
    &Direction&Description\\
    \midrule
    Tiling Modification& $\{(i, j) | i,j \in [0, num\_iters], i\neq j\}$ & Divide the smallest factor from iteration $i$ and multiply to iteration $j$. \\
    Compute-At Modification & $\{-1, 0, 1\}$ & Move backward, stay or move forward in the list of candidates of the computation location. \\
    Parallel-Loops Modification& $\{-1, 0, 1\}$ & The change of the number of fused loops with parallel execution. \\
    Auto-Unroll Modification& $\{-1, 0, 1\}$ & Move backward, stay or move forward in the list of auto-unroll depths. \\
  \bottomrule
\end{tabular}
\end{table*}

Before introducing the \textit{low-level} parameter modification selection process, this section focuses on the construction of its search space.
Taking $1024 \times 1024 \times 1024$ GEMM as an example, with 4 tiling levels, there are 286 tiling choices for each dimension\footnote{since $1024=2^{10}$, the loop tiling can be treated as allocating a series of 10 factors \emph{2} to 4 groups, which is equivalent to find all possible 3-border combinations from 11 positions with repetitions, where ${11+3-1 \choose 3 }= 286$.}.
In the settings where 2 loop unrolling depths are considered and at most 4 levels of spatial loops are fused for thread level parallelism, there are $286 \times 286 \times 286 \times 2 \times 4 \approx 180$ millions of configurations.
In HARL, these configurations are not fully connected, i.e. the RL agent modifies one configuration (or state in the RL context) to construct a \emph{limited} set of nearby configurations, in order to make the gradients of the learning process steady.
The new configuration construction process is an integration of different modification types, which are also known as action subspaces in the RL context. All of them are described in Table~\ref{tab:mod_kind}, and further explained in the following:

\textbf{Tiling Modification}: Computation-intensive operators like matrix multiplications require approporiate tile size for the best cache efficiency. The optimal tile sizes are difficult to be determined by heuristics and are therefore tuned using auto-schedulers.
Suppose that there are $num\_iters$ tiling loops for the considered tensor operator, the actor network consumes the schedule of the current step and outputs an action in the form of a pair of iteration indices $(i, j)$, where $i,j \in [0, num\_iters]$ and $i\ne j$.
$i$ represents the source index, where a minimal factor (greater than 1) will be divided from and multiplied to the target index $j$.
An additional dummy action $(-, -)$ is added to the set of directions to represent that no modification is performed on the current tile sizes.
This process is able to cover the whole search space of tile sizes.

\textbf{Compute-At Modification}: 
Some operations in the subgraph can be computed inline with one of their data producers to improve the data reuse.
However, it is non-trivial to decide which anscestor of all the data producers and which inner loop of the data producer should the operation be computed at.
To implement the compute-at modification for the RL agent, HARL extracts all possible compute-at position pairs $(stage\_id, iter\_id)$,  where $stage\_id$ means the data producer index of the operator, and $iter\_id$ represents the loop index inside the specified data producer.
The compute-at pairs are sorted in sequential order into a candidate list.
The action $a$ is defined to be in the set $\{-1, 0, 1\}$, which means how the compute-at position index should be changed in the candidate list.

\textbf{Parallel-Loops Modification}: 
The number of parallelism of a tiled operation can be modified by changing the numer of fused outer loops for parallel executions.
In this case, the action $a$ is defined in the set $\{-1, 0, 1\}$ for indicating how the number of fused loops should be changed in this modification.

\textbf{Auto-Unroll Modification}: 
The depths of the loop auto-unroll can be controlled by the pragma $unroll$. A list of unroll factors is predefined.
The action $a \in \{-1, 0, 1\}$ indicates the change of index to the next unroll factor in this list.

\subsection{Actor-Critic Based RL for Parameter Modifications}
As described in Section~\ref{sec:overview}, after confirming the optimizing \textit{subgraph} $n$ and its \textit{sketch} $u$ with the help of the MAB-based RL, actor-critic based RL parameter search is conducted in HARL.

In details, actor-critic method~\cite{mnih2016asynchronous} is applied in HARL.
As the name suggests, there are an \textit{actor network} and a \textit{critic network}. The actor network plays a role as an agent. It takes the current state (i.e. \emph{schedule}) as an observation and chooses a set of actions (i.e. \emph{parameter modifications}) to apply to the state for generating the new state as described in Section~\ref{ssec:search_space}. The critic network evaluates the action taken by the actor network and participates in its training process.
The optimization of the actor-critic network follows the policy gradient theorem~\cite{sutton1999policy}. The gradient can be expressed as:
\begin{equation}
\label{eq:policy_gradient}
\nabla_\theta J(\theta) = \mathbb{E}[\Sigma_{t=0}^{T-1} \nabla_\theta \log \pi_\theta(s_t, a_t) A_{\pi_\theta}(s_t, a_t)] 
\end{equation}
In the formula, $t$ represents the current step and $T$ represents the total number of steps.
$s_t$ and $a_t$ are the current state and the chosen action respectively.
$\pi_\theta$ represents the current policy, where the probability of choosing action $a_t$ given state $s_t$ is its output.
$A_{\pi_\theta}$ is the advantage function (a.k.a., the temporal difference error).
\begin{equation}
\label{eq:advantage}
A_{\pi_\theta} = r(s_t, a_t) + \gamma V_{\pi_\theta}(s_{t+1}) - V_{\pi_\theta}(s_t)
\end{equation}
where the function $r(\cdot, \cdot)$ is the reward function, $\gamma$ is the discount factor, and $V_{\pi_\theta}(\cdot)$ is the value network.

\begin{figure}[h]
  \centering
  \includegraphics[width=\linewidth]{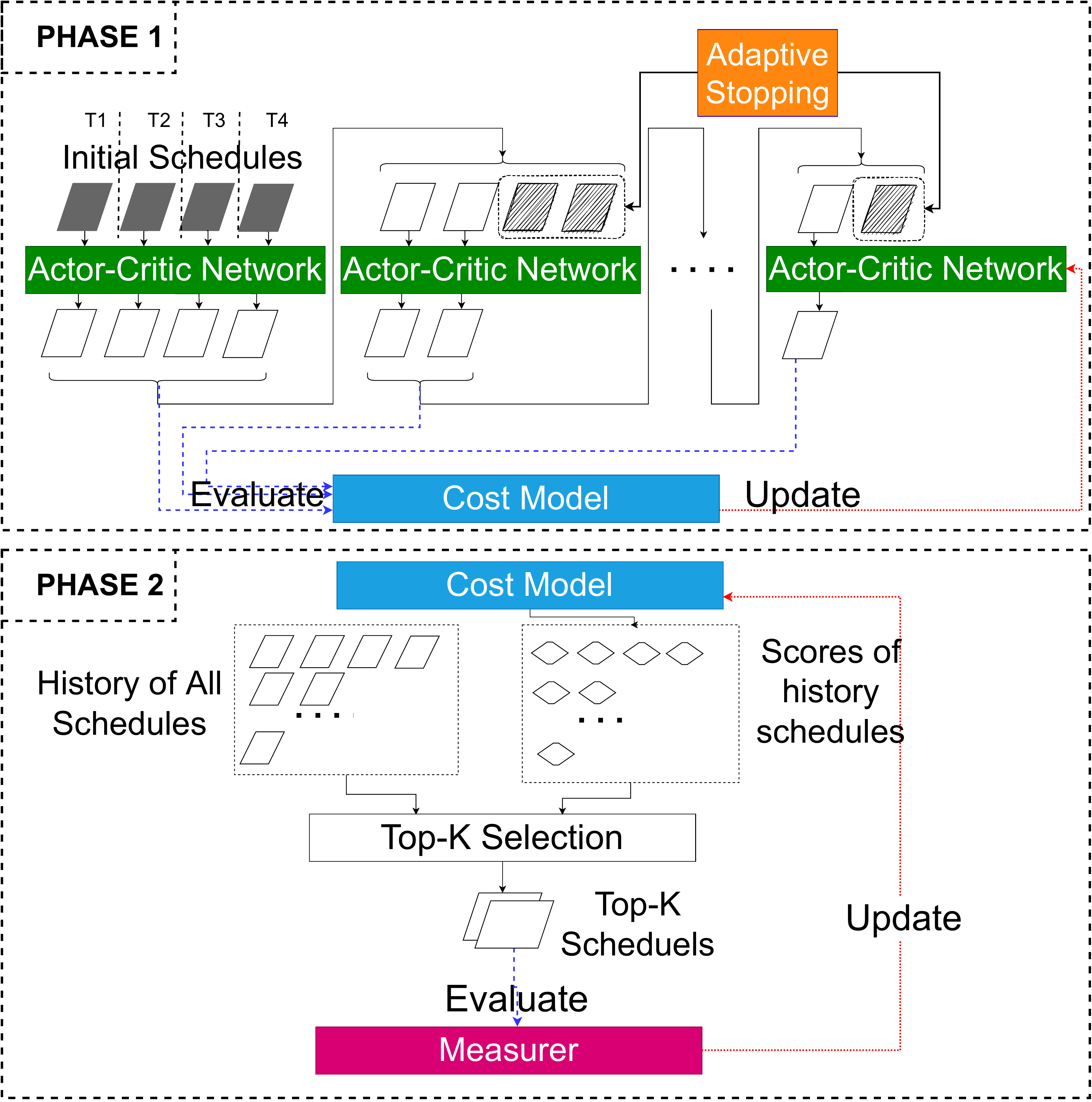}
  \caption{Example of the parameter tuning in HARL}
  \label{fig:param_tune}
  \Description{Example of the parameter tuning in HARL}
\end{figure}

In addition to the actor-critic network, the search process utilizes a light-weight cost model as the reward function $r(\cdot, \cdot)$ to save the time cost by the actual measurements.
There is also an adaptive stopping module which adaptively controls the search process for better search efficiency.
To make the interactions between \textit{actor-critic network}, \textit{adaptive stopping} module, \textit{cost model} and \textit{measurer} clearer, the formualted algorithm of parameter search is shown in Algorithm~\ref{algo:param_search}, and an example of parameter search workflow is demonstrated in Figure~\ref{fig:param_tune}.
The search process is composed of two phases: \textit{parameter modification phase} and \textit{top-k selection phase}.
Following are the detailed descriptions about the two phases.

\textbf{Parameter Modfication Phase}:
This phase is shown as \textbf{PHASE 1} at the top of Figure~\ref{fig:param_tune}.
It maps to an "episode" in the RL language.
At the beginning of this phase, a total of $I$ initial schedules or states $\{s^n_{i, 1}\}_{1 \le i < I}$ are first sampled by randomly filling the sketch $u$ of the subgraph $n$ selected from the process described in Section~\ref{ssec:bandit}.
Each of these initial states undergoes the parameter modifications independently, so we call the path starting from each initial schedule as a \textit{schedule track}.
In Figure~\ref{fig:param_tune}, $I=4$, there are thus 4 initial schedules (shown as gray parallelograms) leading 4 schedule tracks: T1$\sim$T4. The pseudo-code is shown on lines 3-5 in Algorithm~\ref{algo:param_search}.

After passing the states to the actor network, the potentially most beneficial \textit{parameter modification} for each \textit{modification type} is returned and performed on the current states by the agent.
Since each modification type has a \textit{dummy action} that keeps the input state unchanged, the \textit{modifcation type} selection is implicitly performed by the actor network at each step.
Next, the newly generated schedules are passed to the critic network for evaluation using the advantage function $A_{\pi_\theta}$ as defined in Equation~\ref{eq:advantage}.
The advantage values are passed to the \textit{adaptive stopping} module which determines whether to stop searching on a track-wise granularity.
This module controls the time-allocation $\{T^n_i\}_{1 \le i < I}$ for each schedule track $i$.
The \textit{sketched parallelograms} shown in Figure~\ref{fig:param_tune} represents the adaptively stopped schedule tracks.
The details of adaptive stopping are shown in Section~\ref{sec:adaptive_search}.
Meanwhile, the rewards $r(s^n_{i, t}, s^n_{i, t-1}) = \frac{\mathcal{C}(s^n_{i, t}) - \mathcal{C}(s^n_{i, t-1})}{\mathcal{C}(s^n_{i, t-1})}$ are estimated using the cost model $\mathcal{C}$.
The episode goes on as long as there are no less than a minimal number of tracks left.
The above procedure is shown on lines 6-19 in Algorithm~\ref{algo:param_search}.
The states, actions, advantage values and rewards are recorded and used for training the actor and critic network for every specific number of steps.

\textbf{Top-$K$ Selection Phase}: 
This phase is shown as \textbf{PHASE 2} on the bottom of Figure~\ref{fig:param_tune}.
After an episode of parameter modifications finished, \textit{all} travered schedules and their predicted scores are recorded.
In the figure, the gathered parallelograms and diamonds are the collections of visited schedules and their corresponding scores predicted by the cost model respectively.
Next, in order to save the total tuning time from an excessive number of measurements, only the top-$K$ schedules with the best scores are selected for measuring on the target hardware platform.
The measurement results are used to update the cost model $\mathcal{C}$ on the fly.
This top-$K$ selection phase is shown on lines 20-22 in Algorithm~\ref{algo:param_search}.

\begin{algorithm}
\DontPrintSemicolon
\KwIn{Selected Sketch $u$}
\KwOut{Chosen Schedules for Measurements $\hat{S}$, Measured Performance $P$}
\SetKwBlock{Begin}{function}{end function}
\Begin($\text{parameterSearch} {(} u {)}$)
{
  \textbf{Load} $\text{cost model } \mathcal{C} \text{, actor network } \pi_\theta \text{ and value network } V_{\pi_\theta}$\;
  \textbf{Initialize}  number of tracks each round $p$, minimal number of tracks $\hat{p}$\, RL training interval $T_{rl}$\;
  \textbf{Initialize} step $t \gets 0$\;
  \textbf{Initialize} schedules of current round $S$ is sampled from parameter space of sketch $u$ with size $p$\;
  \While{$|S| \ge \hat{p}$}
  {
    Select the set of modification actions $M$ for schedules $S$ using policy $\pi_\theta(M|S)$\;
    Apply each sub-actions in $M$ to the states $S$, and generates new states $S'$\;
    Get reward $R=\frac{\mathcal{C}(S')-\mathcal{C}(S)}{\mathcal{C}(S)}$ \;
    Calculate $Y$ using $A_{\pi_\theta}(S', M, S)$ from Equation~\ref{eq:advantage}\;
    Pass $Y$ to \textit{adaptive-stopping} module to test termination for each schedule.\;
    Record $(S, M, S', R, Y)$ in to the replay buffer $B$\;
    Record $S'$ into the heap $H$ ordered by $\mathcal{C}(S')$.\;
    
    \If{$t \mod T_{rl} = 0$ }
    {
    Sample a mini-batch from the replay buffer $B$\;
    Train actor network $\pi_\theta$ using gradient $\nabla_\theta J(\theta)$ from Equation~\ref{eq:policy_gradient}\;
    Train critic network $V_{\pi_\theta}$ using MSE of the advantage $Y$.\;
    }
    $S \gets S'$\;
    $t \gets t+1$\;
  }\label{endwhile}

  Select top-$K$ states $\hat{S}$ from the heap $H$.\;
  Measure $K$ states and record the performance $(\hat{S}, P)$\;
  Train the cost model $\mathcal{C}(\cdot)$ with $(\hat{S}, P)$\;
    
  \Return{$\hat{S}$, $P$}
}
\caption{Parameter Search Algorithm}\label{algo:param_search}
\end{algorithm}

\section{Apaptive-Stopping Search}
\label{sec:adaptive_search}
Unlike RL scenarios such as game simulations, where an episode ends in specific states, schedule search does not have a well-defined ending state.
In other words, each point accessed during one episode of schedule explorations can be treated as an endpoint and should participate in the top-$K$ selection process at the end of the episode.
In this case, the length control of the schedule tracks is non-trival.
If the length of the schedule track is not enough, the performance of the modified schedules may not be fully exploited.
On the other hand, if the length is excessively large, it would affect the total tuning time, since all tracks of the schedules share the same length no matter if they already hit the maximal performance at some early steps.
Instead of fixing the lengths of all the schedule tracks, HARL uses an adaptive method to determine the lengths on a track-wise granularity.
Meanwhile, it maintains a similar number of candidates as the fixed-length search for the final top-$K$ selection process.

\begin{figure}[h]
  \centering
  \includegraphics[width=\linewidth]{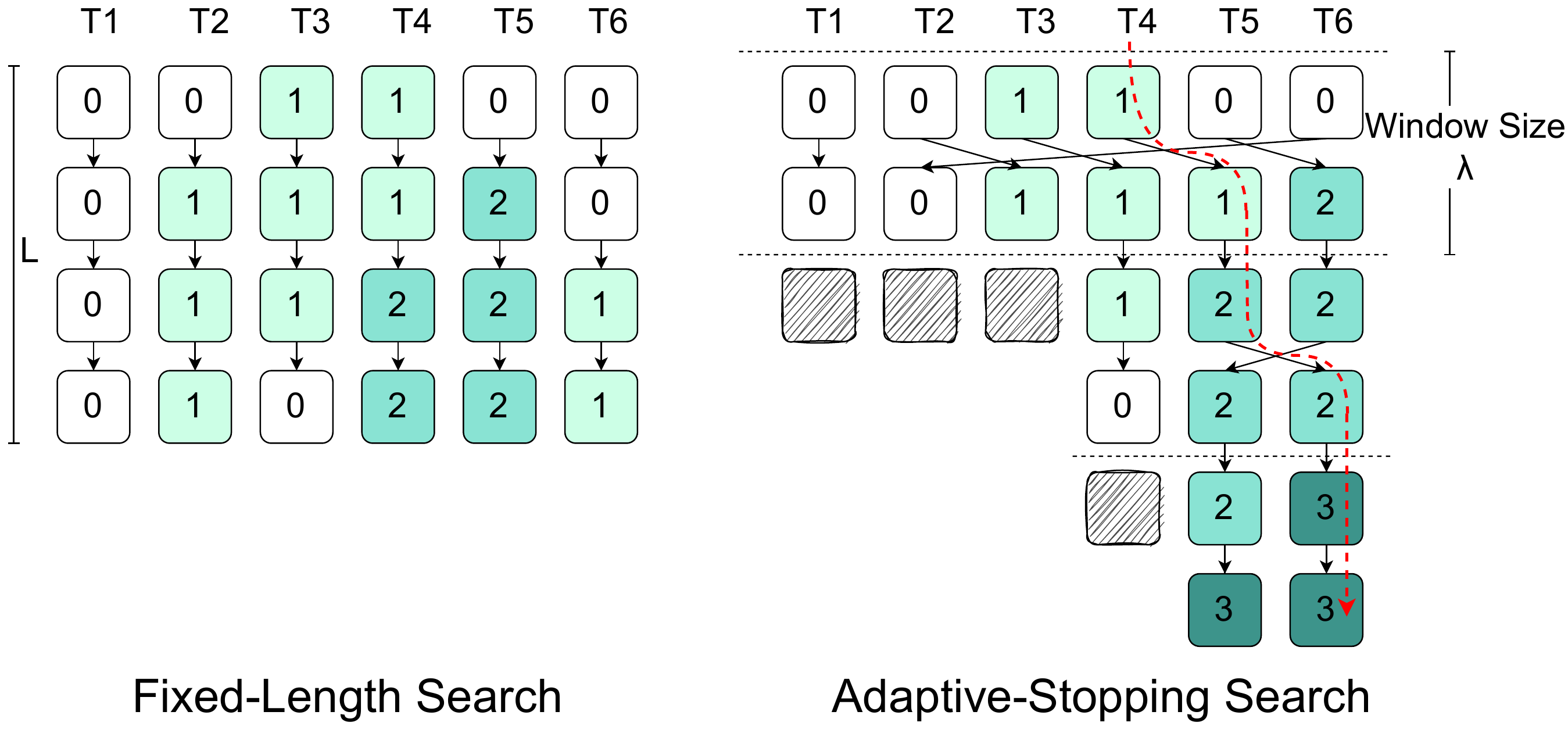}
  \caption{Demo the difference between the fixed-length search and adaptive-stopping search.
    Each box represent a schedule.
    Larger number means better performance.
    Sketched boxes means the eliminated schedules.}
  \label{fig:cmp_search}
  \Description{Adaptive stopping search}
\end{figure}

Figure~\ref{fig:cmp_search} elaborates on the differences between the fixed-length search and the proposed adaptive-stopping search.
Each box in the figure represents a visited schedule and the higher the number inside indicates the better performance.
There are 6 tracks named T1 $\sim$ T6 in this example.
In the \textit{fixed-length} search, as described on the left side of Figure~\ref{fig:cmp_search}, the algorithm searches each of the schedule tracks for $L$ rounds from top to bottom.
The drawback of the fixed-length episodes is that, although schedule tracks T1, T2 and T3 may already reach the maximal performance in the first two rounds, they are still retained for the following modification steps.

For \textit{adaptive-stopping} search, a window size $\lambda$ is defined as a fixed number of steps at which a normal, fixed-length search is performed.
After $\lambda$ steps, the modified schedules are sorted according to the value of \textit{advantages} $A_{\pi_\theta}(\cdot)$ as defined in Equation~\ref{eq:advantage}.
The sorting is represented as the \textit{arrows} between two steps shown in the figure.
The advantage value not only reflects the current performance, but also indicates its expected rewards for further modifications.
A propotion $\rho$ of the schedules with lowest advantages is eliminated, and the search continues for another $\lambda$ rounds. The eliminated tracks are represented in \textit{sketched} boxes.
The elimination-searching process takes place periodically until a minimal number of remaining schedules is met.
In this way, the algorithm tends to explore more on schedules with better potential performance.
In Figure~\ref{fig:cmp_search}, we take T4 as the example:
In the \textit{fixed-length} search, schedule track T4 is explored for $L=4$ rounds, the maximum performance it reaches is $2$.
On the other hand, in the \textit{adaptive-stopping} search, T4 is reprensented by the dotted red track due to the reordering.
Since T4 has a higher rank in the \textit{adavantage} sorting, it is explored for $6$ rounds, thus results in a higher performance $3$.
Meanwhile, if we set $\lambda=L/2$ and $\rho=0.5$ as shown in the right side of the figure, the total number of candidates accessed for the whole episode matches that of the search with fixed length $L$.
By introducing \textit{adaptive-stopping} search, we bring a more efficient search scheme, since at the end of the episode, the candidates for the top-$K$ selection tend to contain better-performed schedules.

\section{Evaluation}
\label{sec:eval}

\subsection{Setup}
To evaluate the effectiveness of HARL on tensor program optimization, a set of experiments consisting of \textit{tensor operator} optimization and \textit{neural network} end-to-end optimization is conducted.
The previous study~\cite{zheng2020ansor} shows that Ansor outperforms Pytorch~\cite{paszke2019pytorch}, AutoTVM~\cite{chen2018learning}, Halide auto-scheduler~\cite{adams2019learning}, and Flextensor~\cite{zheng2020flextensor} on different kinds of tensor programs and hardware platforms, thus we use Ansor (TVM v0.8.0)~\cite{ansorgh} as our only baseline. 
For the actor-critic RL model, we use proximal prolicy optimization (PPO)~\cite{schulman2017proximal} algorithm.
The PPO implementation is adopted from the online source code~\cite{ppopt}.
For the cost model we use Xgboost~\cite{chen2016xgboost} with the same parameters as used in Ansor. More implementation details are shown in Appendix~\ref{app:modelimpl}.
The optimized tensor programs are evaluated on CPU (Intel Xeon 6226R with 32 cores and 2.9GHz) and GPU (Nvidia GeForce RTX 3090). GPU is also used for the RL-based search process.

In the following, we present workloads, settings, metrics and results of two sets of experiments including tensor operator optimization and neural network optimization.

\subsection{Tensor Operator Optimization Comparison}
\label{ssec:exp_op}

\textbf{Workloads}: 
For the tensor operator benchmark, we use GEneral Matrix Multiplication (GEMM)~\cite{gemm} with 3 size categories from small to large: GEMM-S, GEMM-M and GEMM-L.
We also uses convolutions: 1D convolution (C1D), 2D convolution (C2D), 3D convolution (C3D), as well as transposed 2D convolution (T2D)~\cite{dumoulin2016guide}.
The parameters of convolutions are the same as those used in Ansor~\cite{zheng2020ansor}.
Each operator is tested with 4 different sets of parameters.
All the parameters of the tested operators are shown in Appendix~\ref{app:tensorshape}.
These tests are conducted on CPU with $1$ and $16$ batch sizes.

\textbf{Search Settings}:
All tensor operator optimizations are conducted with 1000 measurement trials on both Ansor and HARL for fair comparisons.
Ansor and HARL are setup with same number of measurement candidates in each round of exploration for verifying the efficiency of proposed adaptive-stopping search.
For adaptive stopping, we set the window size $\lambda=20$, and the elimination ratio $\rho=0.5$.
More details on the parameter selections are shown in Appendix~\ref{app:sens}.
For the learning process of actor-critic network, the learning rates are $3e-4$ and $1e-3$ for actor and critic networks respectively.
The discount factor $\gamma$ is set to $0.9$.
In the critic network, the weight of the MSE loss is $0.5$ and that of distribution entropy is $0.01$.
For the sketch MAB, the constant $c$ used in SW-UCB defined in Equation~\ref{eq:sw-ucb} is set to $0.25$ and the window size is $256$.

\begin{figure}[h]
  \centering
  \includegraphics[width=\linewidth]{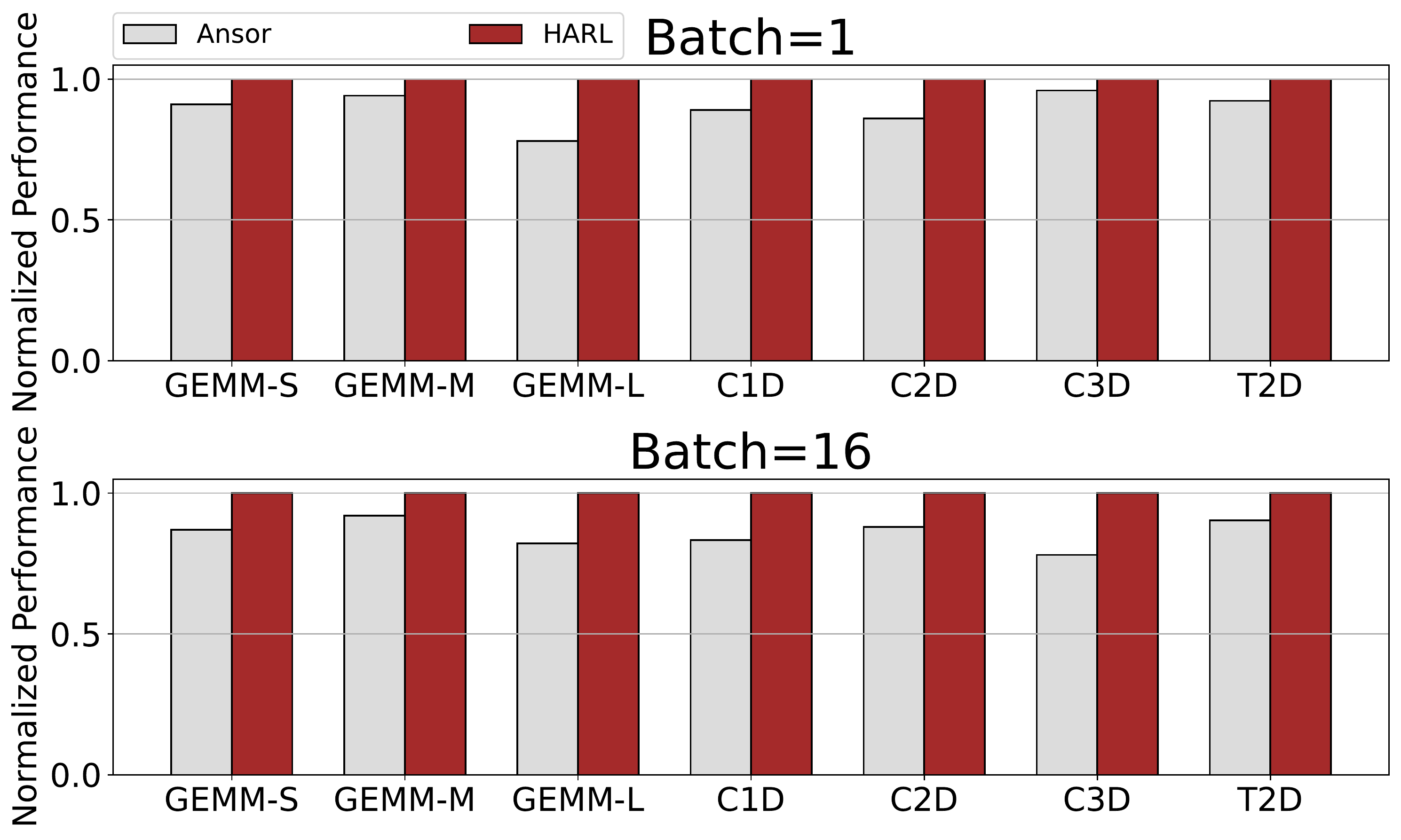}
  \caption{The normalized performance (inverse of execution time) of Ansor and HARL on different tensor operators.}
  \label{fig:exp_op_throughput}
  \Description{experiments on operator performance}
\end{figure}

\begin{figure}[h]
  \centering
  \includegraphics[width=\linewidth]{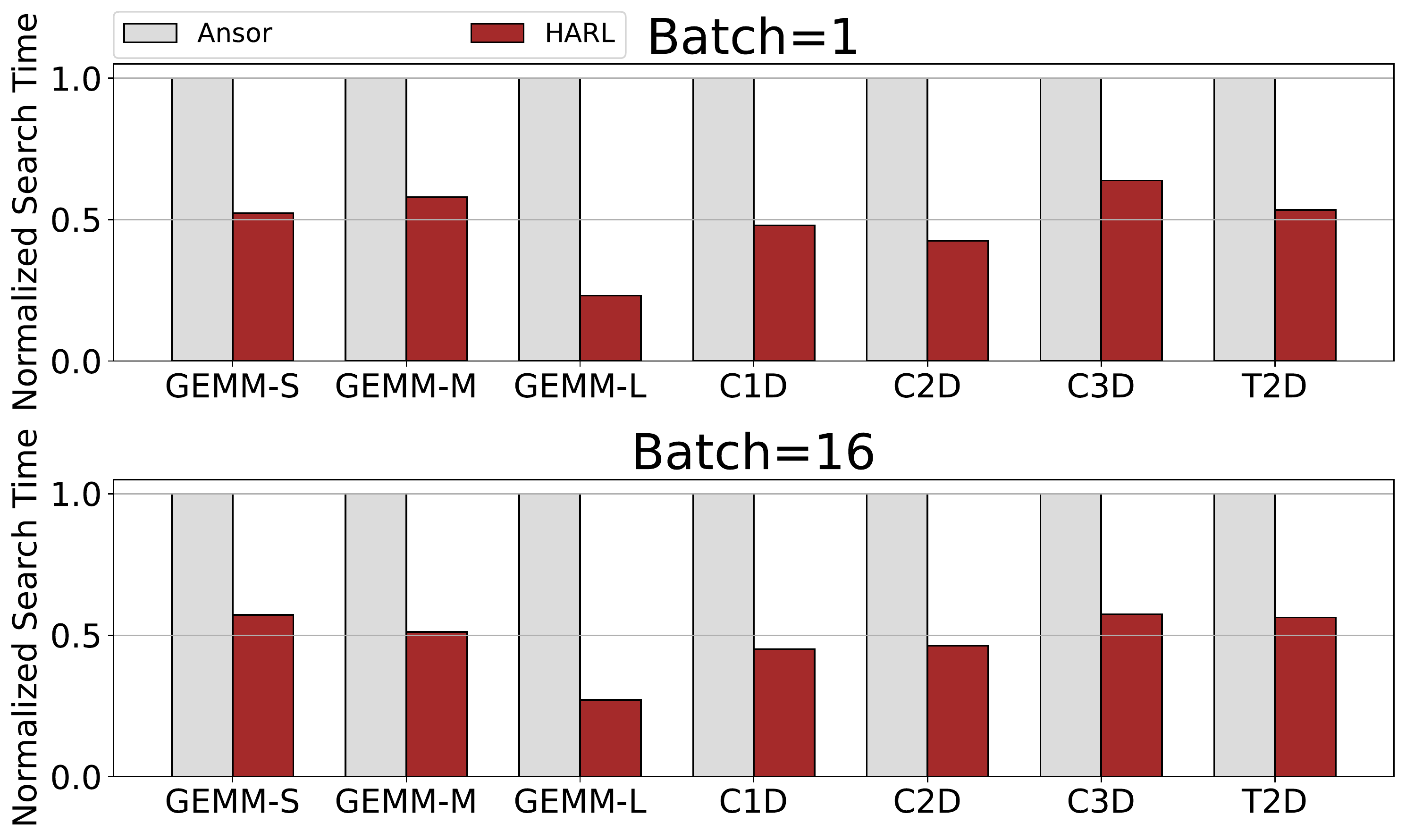}
  \caption{The normalized search time of Ansor and HARL on different tensor operators.}
  \label{fig:exp_op_search_time}
  \Description{experiments on operator search time}
\end{figure}

\textbf{Metrics}:
To verify the effectiveness of HARL on the search speed of tensor programs and the performance of the output program, two metrics are used in the experiments.
\emph{Performance} is the inversed execution time of the output programs produced by the auto-schedulers.
\emph{Search time} is the time consumed by the auto-scheduler to find the tensor program with the performance no worse than the final output program of the baseline.
To make the comparison between baselines and HARL clearer, the \emph{Performance} and \emph{Search time} metrics are normalized to values in the range $[0, 1]$.

\textbf{Results}:
As shown in Figure~\ref{fig:exp_op_throughput}, the output schedules of HARL outperform those of Ansor on every tensor operations by 6\% $\sim$ 22\%.
The search time improvement is more significant, as shown in Figure~\ref{fig:exp_op_search_time}.
On average, to achieve the performance no worse than the best schedule achieved by Ansor, HARL uses 63\% $\sim$ 23\% search time.
The GEMM-L operation has the most outstanding performance improvement.
This is due to their larger data dimensions, thus they have larger tile size configuration spaces.
The GEMM-L operations with large search space bring harder problems to auto-schedulers, and HARL is able to make profound performance increase with the help of the adaptive length control and hierarchical learning-based search method.

\begin{figure}[h]
  \begin{minipage}[t]{0.99\linewidth}
    \vspace{0pt}
    \centering
    \centerline{\includegraphics[width=7.5cm]{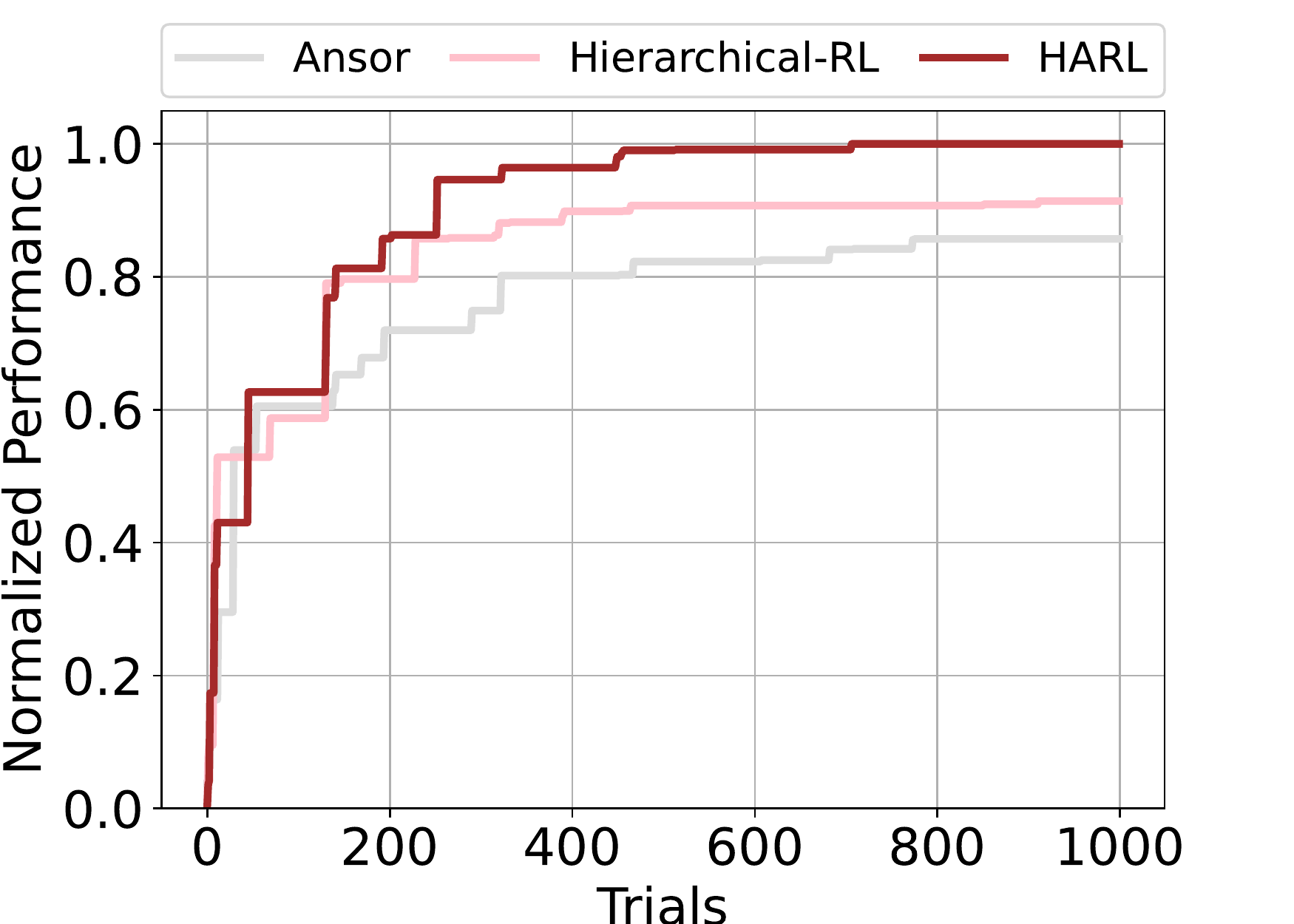}}
    \begin{center}{(a) The ablation study of hierarchical RL and adaptive-stopping .}\end{center}\medskip
  \end{minipage}
  \begin{minipage}[t]{0.99\linewidth}
    \vspace{0pt}
    \centering
    \centerline{\includegraphics[width=7.5cm]{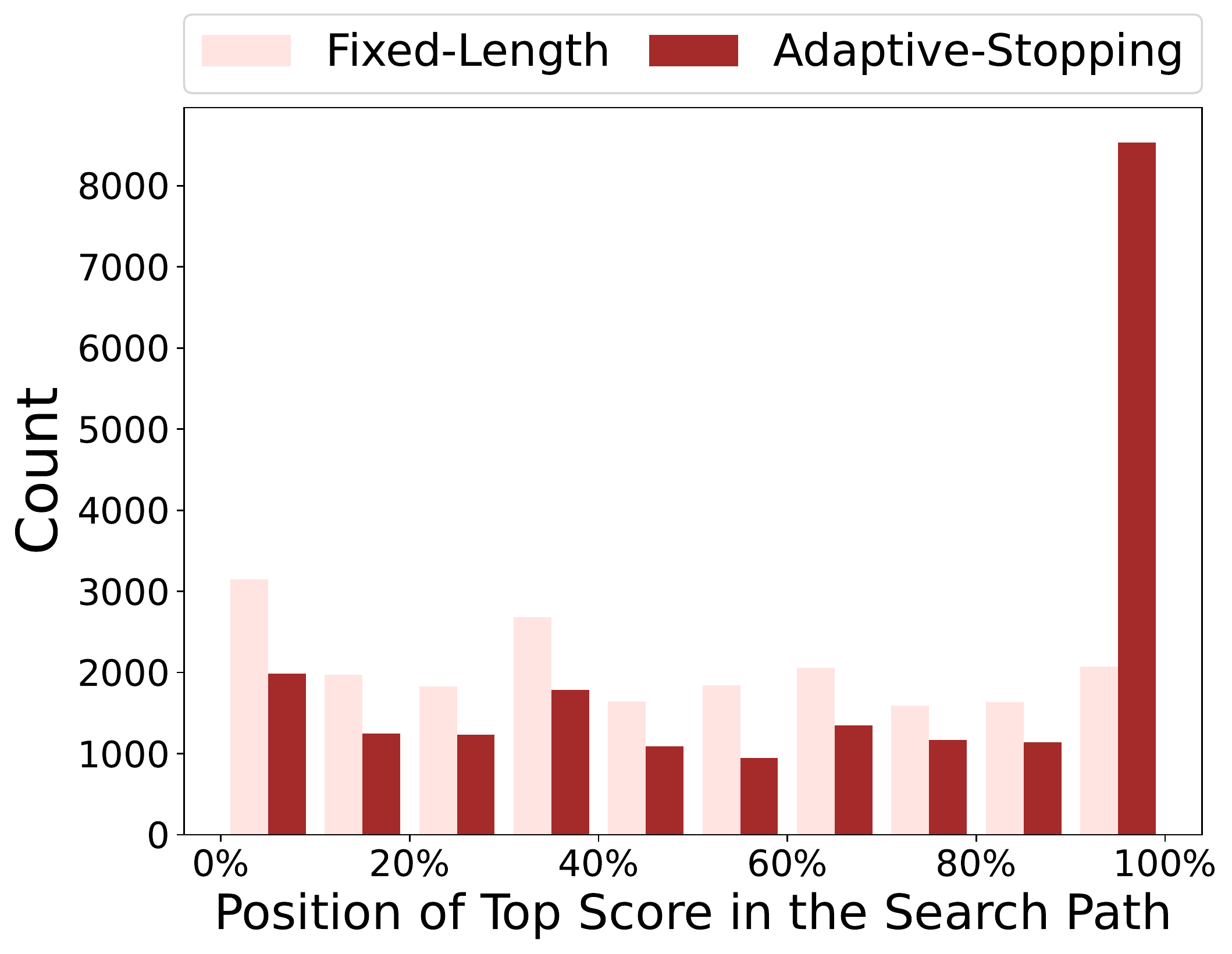}}
    \begin{center}{(b) Histogram comparison of fixed-lengths vs. adaptive-stopping on the schedule tracks efficiency. X-axis represents the position of the best-performed schedule (i.e. \emph{critical step}) on the search path divided by the path length. Y-axis is the frequency of these relative positions collected during auto-tuning.}\end{center}\medskip
  \end{minipage}
  
  \caption{Ablation Studies}
  \label{fig:exp_op_ablation}
  \Description{ablation study }
\end{figure}

\begin{figure}[h]
  \centering
  \includegraphics[width=\linewidth]{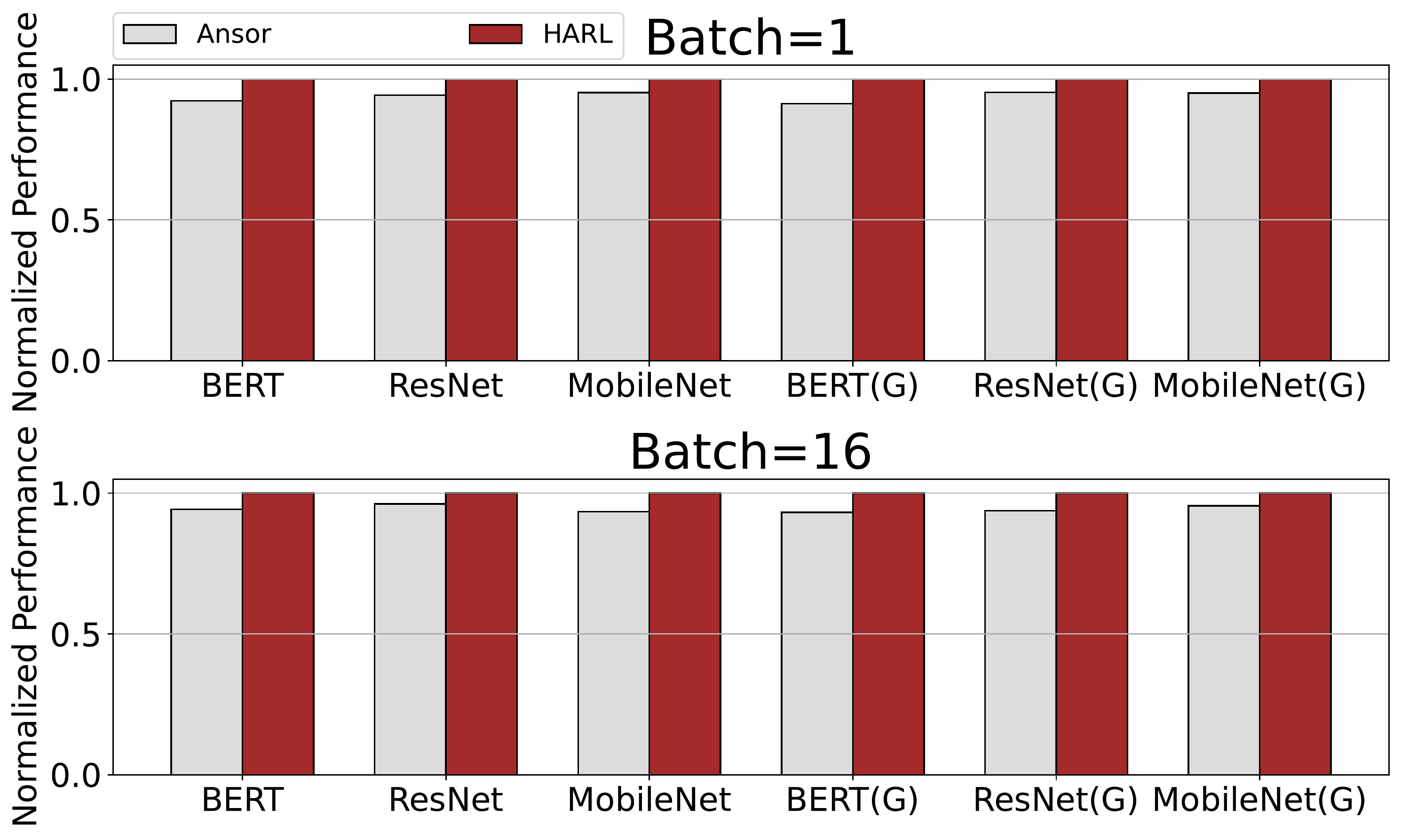}
  \caption{The normalized performance of Ansor and HARL on different neural networks. suffix (G) means GPU tests.}
  \label{fig:exp_nn_throughput}
  \Description{experiments on NN performance}
\end{figure}

\begin{figure}[h]
  \centering
  \includegraphics[width=\linewidth]{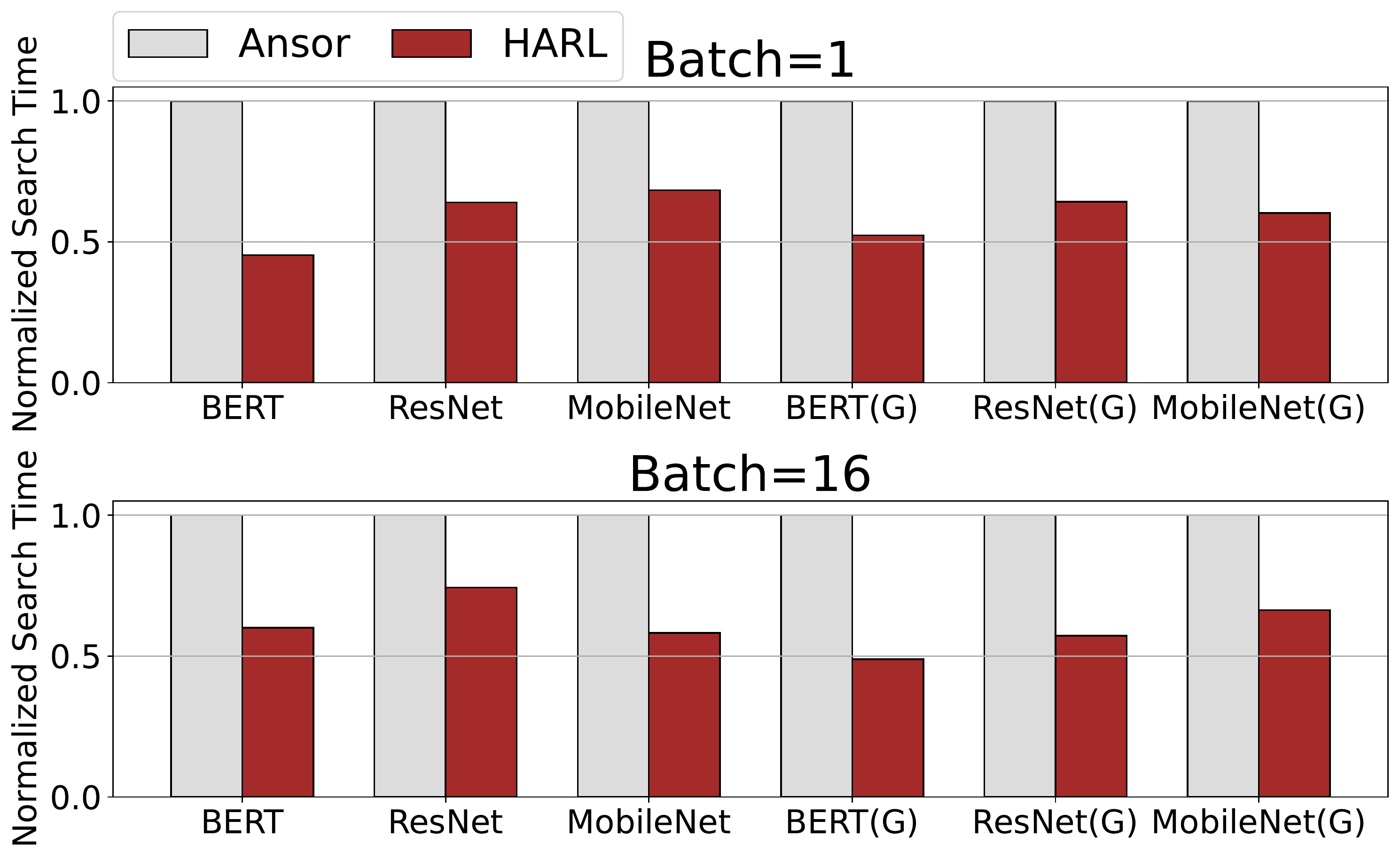}
  \caption{The normalized search time of Ansor and HARL on different neural networks. suffix (G) means GPU tests.}
  \label{fig:exp_nn_search_time}
  \Description{experiments on NN search time}
\end{figure}
\textbf{Ablation Study}:
Figure~\ref{fig:exp_op_ablation}(a) shows the records when optimizing one GEMM-L operation on \textit{Ansor}, \textit{HARL with fixed-length episodes}, and \textit{HARL with adaptive-stopping modules}.
HARL is able to outperform Ansor at early steps.
"Hierarchical-RL" in the figure means the HARL implementation with fixed-length search.
From the figure, the \textit{adaptive-stopping} module is able to significantly improve the search efficiency of HARL after the optimization goes for a few hundreds of rounds.
This is because \textit{adaptive-stopping} relies on the accuracy of the \textit{adavantage} function $A_{\pi_\theta}$.
This function is able to accurately predict the advantage value of the agent's movement only when the critic network and the cost model are sufficiently trained.
Without the help of \textit{adaptive-stopping}, \textit{Hierarchical-RL} is still able to outperform Ansor.
This shows the effectiveness of non-stationary MAB modeling on sketch selection.

To further examine the search efficiency of the \emph{adaptive-stopping} module, 
we record the step where the highest performance score is reached in each schedule track during the search process.
We call the top-scored step as the \emph{critical step}.
All the steps after this \emph{critical step} do not find a better schedule, thus they are called \emph{wasted steps}.
The historgrams of the \emph{critical steps} for the two search methods are shown in Figure~\ref{fig:exp_op_ablation}(b).
In \emph{adaptive-stopping}, most of the \emph{critical steps} appears in the last 10\% steps of their schedule tracks, which indicates less than 10\% \emph{wasted steps}.
Additionally, 16\% of these \emph{low-waste} schedule tracks are from the longest tracks of \emph{adaptive-stopping}, which is not shown in the figure.
The smaller number of \emph{wasted steps} in \emph{adaptive-stopping} leads to a more efficient search process.

\subsection{Neural Network Optimization Comparison}
\label{ssec:exp_nn}

\textbf{Workloads}: 
For end-to-end neural network optimizations, we test Ansor and HARL on ResNet50~\cite{he2016deep}, MobileNet-V2~\cite{sandler2018mobilenetv2}, and BERT~\cite{devlin2018bert}. They are commonly used DNN models utilized in computer vision (CV), and natural langurage processing (NLP).
They are tested on Intel CPU and Nvidia GPU with batch size $1$ and $16$.

\textbf{Settings}: 
The number of trials measured is 12,000 for BERT, 22,000 for ResNet50, and 16,000 for MobileNet-V2.
For the actor-critic network and MAB settings, the parameters used in end-to-end benchmarks are the same as decribed in Section~\ref{ssec:exp_op}.

\begin{table}
  \caption{Ablation Study for BERT on CPU}
  \label{tab:bert_ablation}
  \begin{tabular}{p{4cm}p{2cm}c}
    \toprule
    &Execution Time Contribution&SpeedUp\\
    \midrule
    GEMM-I & 25.3\% & 1.15x \\
    GEMM-IV & 25.1\% & 1.10x \\
    GEMM-III & 20.5\% & 1.06x \\
    GEMM-II & 11.4\% & 1.09x \\
    Softmax & 5.0\% & 1.14x \\ [10pt]
    Batch\_GEMM-II & 4.0\% & 0.98x \\
    Batch\_GEMM-I & 3.8\% & 1.01x \\
    Element-wise-II & 2.4\% & 0.95x \\
    Element-wise-I & 2.2\% & 1.01x \\
    GEMM+Tanh & 0.4\% & 1.02x \\
    \midrule
    Estimated HARL (sum) & 100\% & 1.10x \\
    Measured HARL & - & 1.08x \\
    Measured HARL (w/o subgraph MAB) & - & 1.06x \\
    
  \bottomrule
\end{tabular}
\end{table}

\textbf{Results}: 
Figure~\ref{fig:exp_nn_throughput} shows the performance comparisons on different neural networks and hardware platforms.
Figure~\ref{fig:exp_nn_search_time} shows the comparisons on the search time.
HARL improves the performance of the outcome by 8\% and 9\% for CPU and GPU respectively.
The search time reduction is up to 55\% and 51\%.

Table~\ref{tab:bert_ablation} further dives into the execution time contributions of different subgraphs.
In the table, the first column contains the 10 subgraphs of the BERT model.
The second column is the percentage of each subgraph's contribution to the total execution time of HARL's output program.
The third column shows the speedup of HARL relative to Ansor on each subgraph.
The first 5 subgraphs are with large number of floating-point operations and contribute to 87\% of the total execution time. 
Their speedup matches that of the individual tensor operators as evaluated in Section~\ref{ssec:exp_op}.
For the remaining subgraphs, they are with smaller execution time contributions.
For example, the \textit{Batch GEMMs} are with magnitudes of smaller number of floating operations ($\sim$ 1/3000) than that of GEMM operations.
Taken together, they take 13\% of the total execution time, and HARL has similar performance as Ansor on them.
Note that, the subgraphs communication overheads affects the end-to-end performance as shown in comparison of \emph{Estimated HARL} and the \emph{Measured HARL} in the table.
Nonetheless, HARL still achieves the significant performance improvement and search time reduction with the help of hierarchical RL and adaptive-stopping technique.

\begin{figure}[htb]
  \centering
  \includegraphics[width=\linewidth]{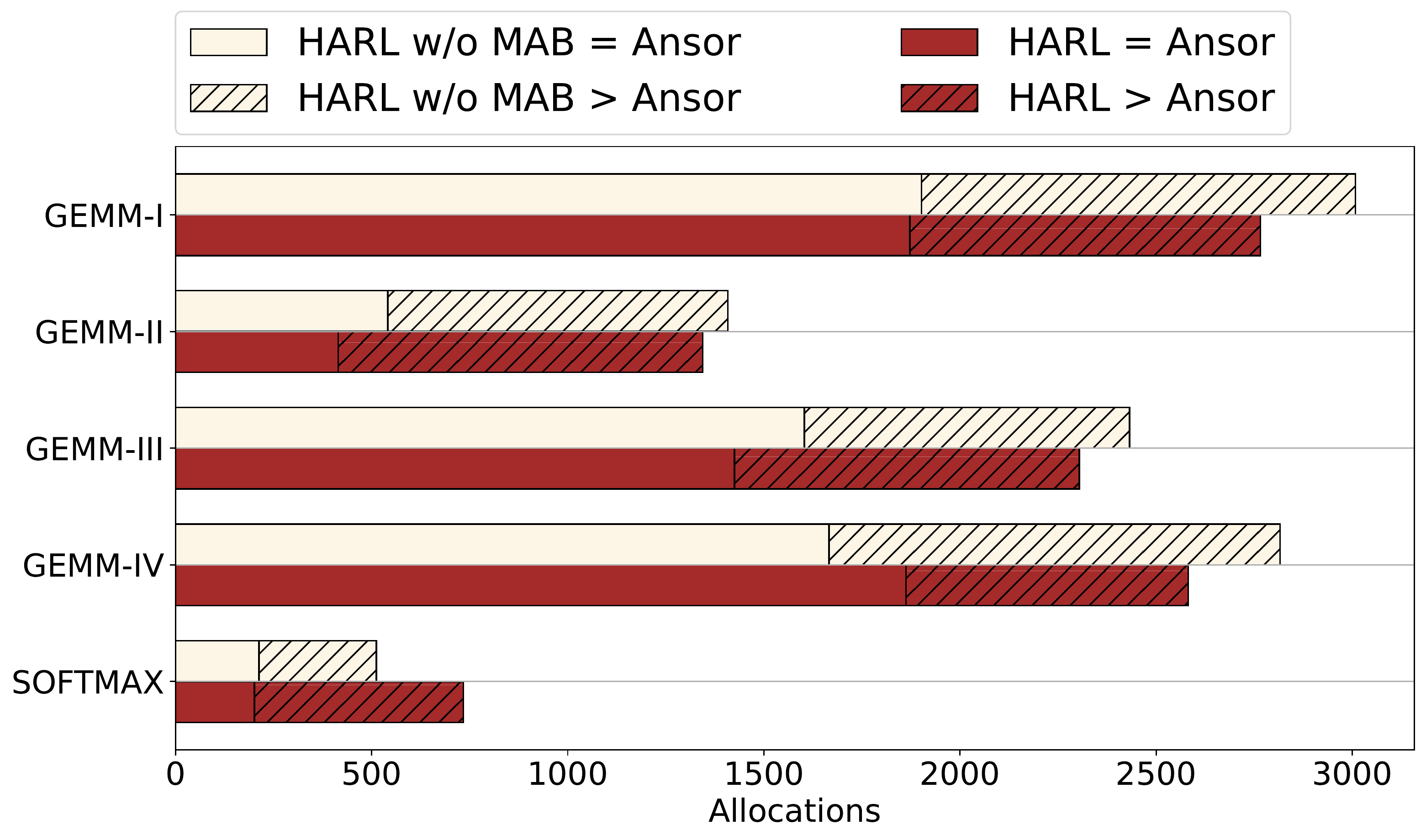}
  \caption{The ablation study of the subgraph MAB algorithm on the subgraph trial-allocations. The ’= Ansor’ part means the trials allocated at the time
the system reaches the best schedule of Ansor. The ’> Ansor’ part (sketched bars)
represents the remaing trials allocated after reaching Ansor’s best
performance.}
  \label{fig:mab_ablation}
  \Description{ablation study on subgraph MAB}
\end{figure}

To analyze the effectiveness of subgraph MAB selection that is only applicable in the end-to-end neural network optimization, the test result on \emph{HARL without subgraph MAB} is presented at the bottom of Table~\ref{tab:bert_ablation}.
It shows that without the subgraph MAB selection process, HARL has a worse performance due to the greedy subgraph time-allocations similar as Ansor.

To better analyze this phenomenon, we list the subgraph trial-allocation details $\{T^n\}_{1 \le n < N}$ in Figure~\ref{fig:mab_ablation}.
The experiments are performed on \emph{HARL} and \emph{HARL without subgraph MAB modeling}.
The trial-allocations are split into two parts. 
The \emph{'= Ansor'} part means the trials allocated at the time the system reaches the best schedule of Ansor.
The \emph{'> Ansor'} part represents the remaing trials allocated after reaching Ansor's best performance.
When looking at the \emph{total} allocated trials, all the time-consuming GEMM subgraphs in \textit{HARL} with MAB (dark-red bars) have fewer allocations than those in \textit{HARL without subgraph MAB}.
Thus, the subgraph MAB algorithm successfully avoids the over-allocation on subgraphs with high early-trials improvements.
Instead, the trials allocated for the \textit{SOFTMAX} subgraph are improved, which leads to a high speed-up (1.14x) as shown in Table~\ref{tab:bert_ablation}.
On the other hand, when looking at the part of the trials allocated when the systems just reach Ansor's best performance, i.e. the bars that are not sketched, the similar improvement is observed.
When using subgraph MAB method, the trials allocated for GEMMs with lower total improvements \emph{GEMM-II} and \emph{GEMM-III} are fewer than those in the greedy subgraph selection approach.
Instead, more trials are allocated for \emph{GEMM-IV} which has more potential for improvement, as indicated in Table~\ref{tab:bert_ablation}.

\section{Conclusion}
\label{sec:conclusion}
This work proposes HARL, a hierarchical and adaptive reinforcement learning-based auto-scheduler for tensor programs.
It is template-free with improved performance on the generated codes and significantly faster tuning process.
The effectiveness of HARL is verified on different kinds of tensor operators and DNNs, under CPU and GPU hardware platforms.
We believe that, HARL is an innovation on how to utilize the advanced reinforcement learning based decision process to facilitate the time-consuming tensor code optimization.

\begin{acks}
This work is in part supported by Centre for Trusted Internet and Community, National University of Singapore.
\end{acks}

\bibliographystyle{ACM-Reference-Format}
\bibliography{harl}

\appendix

\section{Artifact Descriptions}

\begin{table*}[t]
  \caption{Model Parameters}
  \label{tab:params}
  \begin{tabular}{rcl}
    \toprule
    &Default Value&Description\\
    \midrule
    $\lambda$ & 20 & adaptive-stopping window size \\
    $\rho$ & 0.5 & adaptive-stopping elimination rate \\
    $\hat{p}$ & 64 & adaptive-stopping minimum number of remaining tracks\\
    $lr_a$ & 3e-4 & learning rate of actor network \\
    $lr_c$ & 1e-3 & learning rate of critic network \\
    $T_{rl}$ & 2 & actor-critic network training interval\\
    $\gamma$ & 0.9 & discount factor in Equation~\ref{eq:advantage} \\
    $w_{MSE}$ & 0.5 & critic network MSE loss weight\\
    $w_{entropy}$ & 0.01 & entropy loss weight\\
    $c$ & 0.25 & SW-UCB constant in Equation~\ref{eq:sw-ucb}\\
    $\tau$ & 256 & SW-UCB window size \\
    $\alpha$ & 0.2 & historical gradient importance in Equation~\ref{eq:ansor_task}\\
    $\beta$ & 2 & similar subgraph importance in Equation~\ref{eq:ansor_task}\\
    $r_{min}$& 1 & min number of seconds for repeated measurement of a schedule \\
  \bottomrule
\end{tabular}
\end{table*}

\begin{table}
  \caption{Tensor Operator Configurations}
  \label{tab:op_config}
  \begin{tabular}{cp{2cm}p{3cm}}
    \toprule
    Operator&Description&Configurations\\
    \midrule
    GEMM-S & $(M, K, N)$ & (128,128,128), (128,256,128), (256,256,256), (512,32,512) \\
    GEMM-M & $(M, K, N)$ & (512,512,512), (128,1536,512), (128,512,1536), (256,1024,512) \\
    GEMM-L & $(M, K, N)$ & (1024,1024,1024), (128,3072,768), (128,768,3072), (256,1536,768) \\
    C1D & $(L, C_i, C_o, K,$ $stride, padding)$ & (256,64,128,3,2,1), (128,128,256,1,2,0), (64,256,256,5,1,2), (32,512,512,3,1,1) \\
    C2D&$(H, W, C_i, C_o, K,$ $stride, padding)$& (224,224,3,64,7,2,3), (56,56,64,64,1,1,0), (14,14,256,256,3,1,1), (7,7,512,512,3,1,1) \\
    C3D&$(D, H, W, C_i, C_o$, $K,$ $stride, padding)$& (16,224,224,3,64,7,2,3), (16,56,56,64,64,1,1,0), (16,14,14,256,256,3,1,1), (16,7,7,512,512,3,1,1) \\
    T2D&$(H, W, C_i, C_o, K,$ $stride, padding)$& (4,4,512,256,4,2,1), (8,8,256,128,4,2,1), (16,16,128,64,4,2,1), (32,32,64,3,4,2,1)\\
  \bottomrule
\end{tabular}
\end{table}
\subsection{Model Implementation Details}
\label{app:modelimpl}
The PPO implementation is adopted from the online source code~\cite{ppopt}.
Modifications are made to the actor network in order to produce actions probabilities for multiple parameter modification subspaces as explained in Section~\ref{ssec:search_space}.
The number of actions for each subspace corresponds to the number of directions as shown in Tabel~\ref{tab:mod_kind}.
More specifically, the number of actions is $num\_iters\times num\_iters + 1$ (1 is for the dummy action) for tile size modification and 3 for the other modification types(compute-at, parallel-loops, and autounroll).
For the cost model, we use Xgboost from sklearn python package.
The parameters of the models mentioned in the paper are listed in Table~\ref{tab:params}.

\begin{table}
  \caption{Sensitivity analysis for adaptive-stopping window size $\lambda$ on 1024x1024x1024 GEMM. Each experiment runs for 1000 trials.}
  \label{tab:sens_lambda}
  \begin{tabular}{cp{2cm}p{3cm}}
    \toprule
    $\lambda$&Normalized Performance&Normalized Time/Iteration\\
    10 & 0.917 & 0.491\\
    20 & 0.991 & 0.558\\
    40 & 1.0 & 0.706\\
    80 & 0.993 &1.0\\
  \bottomrule
\end{tabular}
\end{table}

\begin{table}
  \caption{Sensitivity analysis for adaptive-stopping elimination ratio $\rho$ on 1024x1024x1024 GEMM. Each experiment runs for 1000 trials.}
  \label{tab:sens_rho}
  \begin{tabular}{cp{2cm}p{3cm}}
    \toprule
    $\rho$&Normalized Performance&Normalized Time/Iteration\\
    0.75 & 0.864 & 0.748\\
    0.5 & 0.990 & 0.817\\
    0.25 & 1.0 & 1.0 \\
  \bottomrule
\end{tabular}
\end{table}

The depths of \emph{auto-unroll modification} described in Section~\ref{ssec:search_space} are different for CPU and GPU environment:
\begin{itemize}
    \item CPU: 0, 16, 64, 512
    \item GPU: 0, 16, 64, 512, 1024
\end{itemize}

\subsection{Platform Configuration}
\label{app:platform}
The optimized tensor programs are evaluated on both CPU and GPU platforms.
\begin{itemize}
    \item CPU platform is configured with Intel Xeon 6226R with 32 cores and 2.9GHz. 128GB memory is used. The OS is Ubuntu 20.04. AVX512 instruction set is supported.
    \item GPU platform is configured with one Nvidia GeForce RTX 3090.
\end{itemize}

\subsection{Shapes of Tensor Operators}
\label{app:tensorshape}
The different shapes for the tensor operators used in the evaluation in Section~\ref{ssec:exp_op} are shown in Table~\ref{tab:op_config}.
The convolution parameters are the same with those used in Ansor~\cite{zheng2020ansor}.

\subsection{Hyper Parameters Selection}
\label{app:sens}
\textbf{Adaptive-stopping window size $\lambda$}: The window size is set to 20 in the default settings. Experiments on different settings are shown in Table~\ref{tab:sens_lambda}. To make the search time metric directly reflect the running speed instead of the search efficiency, we use the normalized time spent for each search iteration as the metric. As shown in the table, smaller $\lambda$ affects the final performance, since the potentially well-performed schedule tracks are not sufficiently explored before being eliminated. However, a larger $\lambda$ does not always result in an efficient search. As $\lambda$ increases, the improvement of the number of potentially good candidates becomes marginal, since they are selected by approximations instead of the actual measurements. However, the search time is largely affected due to increased episode lengths, and it starts to match the time spent on measurements.

\textbf{Adaptive-stopping elimination rate $\rho$}: Similarly to the window size $\lambda$, the elimination rate also faces the trade-off between the performance and the search time.
We tried 3 different values $0.25, 0.5, 0.75$ on elimination rate $\rho$, and found that $\rho=0.25$ eliminates less candidates but produces tiny performance improvements on $\rho=0.5$, while $\rho=0.75$ has a significant performance drop. For the best search time and performance consideration, we choose $\rho=0.5$ as the default setting of the experiments.

\end{document}